\documentclass[sigconf]{acmart}

\setcopyright{none}

\renewcommand\footnotetextcopyrightpermission[1]{} 
\usepackage{mathtools}
\usepackage{amsmath}
\usepackage{epstopdf}
\usepackage[linesnumbered, ruled,vlined]{algorithm2e}
\SetKwRepeat{Do}{do}{while}%
\SetKw{Continue}{continue}
\SetKw{Break}{break}
\usepackage{graphicx}
\usepackage{subfig}

\def \queryNameCap {Category Aware Multi-criteria route planning}
\def \queryName {category aware multi-criteria route planning}
\def \queryAbb {CAM}

\def \algoName {Global Category Nearest Neighbour}
\def \algoAbb {GCNN}

\newcommand{\argmin}[1]{\underset{#1}{\operatorname{arg}\,\operatorname{min}}\;}

\newtheorem{prune}[]{Pruning Rule}

\begin{document}
\title{An Efficient Approximation Algorithm for Multi-criteria Indoor Route Planning Queries}

\author{Chaluka Salgado}
\orcid{0000-0001-5516-8271}
\affiliation{%
  \institution{Monash University}
  \city{Clayton}
  \state{Australia}
}
\email{chaluka.salgado@monash.edu}

\author{Muhammad Aamir Cheema}
\affiliation{
  \institution{Monash University}
  \city{Clayton}
  \state{Australia}
}
\email{aamir.cheema@monash.edu}

\author{David Taniar}
\affiliation{
  \institution{Monash University}
  \city{Clayton}
  \state{Australia}
}
\email{david.taniar@monash.edu }
\renewcommand{\shortauthors}{Chaluka et al.}

\begin{abstract}
A route planning query has many real-world applications and has been studied extensively in outdoor spaces such as road networks or Euclidean space. Despite its many applications in indoor venues (e.g., shopping centres, libraries, airports), almost all existing studies are specifically designed for outdoor spaces and do not take into account unique properties of the indoor spaces such as hallways, stairs, escalators, rooms etc. We identify this research gap and formally define the problem of category aware multi-criteria route planning query, denoted by CAM, which returns the optimal route from an indoor source point to an indoor target point that passes through at least one indoor point from each given category while minimizing the total cost of the route in terms of travel distance and other relevant attributes. We show that CAM query is NP-hard. Based on a novel \emph{dominance-based} pruning, we propose an efficient algorithm which generates high-quality results. We provide an extensive experimental study conducted on the largest shopping centre in Australia and compare our algorithm with alternative approaches. The experiments demonstrate that our algorithm is highly efficient and produces quality results.
\end{abstract}


\maketitle

\section{Introduction} \label{sec:intro}
People spend a significant amount of their time in indoor spaces often in unfamiliar buildings such as shopping malls, airports, and libraries~\cite{jenkins1992activity}. Recent advances in indoor positioning technology, cheap wireless network and availability of geo-tagged data have resulted in huge demand for
indoor location-based services such as finding nearby indoor objects, indoor navigation, and route planning to name a few.
Route planning is one of the most popular services among both indoor and outdoor users, which assists them in planning a route satisfying their preferences.
Specifically, a user may issue a route planning query by providing a source location and a target location along with her preferences as a set of keywords (e.g., restaurant, salon, supermarket). A route planning query returns an optimal route that starts from the source location, passes through at least one location from each given preference and ends at the target location.

Due to its popularity, route planning query has been extensively studied in the past few years~\cite{li2005trip,sharifzadeh2008optimal,cao2012keyword,zeng2015optimal,yao2011multi}.
However, all these techniques are specifically designed for outdoor spaces and cannot be efficiently extended for the indoor spaces
because they fail to exploit the
unique properties specific to indoor venues. For example, indoor graphs have a much higher out-degree as compared to the road networks~\cite{shao2016vip}. Furthermore, the object density is much higher for indoor venues, e.g., the number of POIs (e.g., restaurants, fuel stations) on the vertices of road networks is typically small whereas the number of objects in a single room (e.g., products in a supermarket) of indoor venues may be in thousands. Thus, specialized techniques
are required to answer route planning queries in indoor venues.

Inspired by the above, in this paper, we provide first set of techniques to answer an important route planning query with various applications in different
scenarios. Consider a user who is in the car park of a large shopping center and has a list of  items to buy (e.g., a wine bottle, a bunch of flowers, a cake, and a wrist watch). She may want to find an optimal route such that the total distance she needs to walk and the total price she pays to purchase all these items are minimized. She may use a \textit{\queryNameCap} query, denoted as \queryAbb, which takes as input a set of categories (e.g., the list of items she wants to purchase) and a scoring function, and returns the route that passes through at least one object of each category and has the minimum score where  score of each route is computed using the user defined scoring function considering the total length of the route and total price of the items along the route.

In contrast to traditional route planning queries that only consider a single criterion (i.e., distance), \textit{\queryNameCap} queries could retrieve optimal route considering multiple criteria such as the total length of the route, total price, total
rating of items, and total waiting time for the activities etc. Consider another example of a user in an airport who is running late for a flight
and needs to withdraw money from an ATM, grab a coffee, and needs to go to a service desk before she checks in. For such a user, total length of
the route is important as well as the total waiting time at the ATM, coffee machine and service desk. Therefore, she may issue a \queryAbb~query
where the scoring function is used to compute the score of a route considering its total length and the total waiting time at each facility (i.e., ATM, coffee machine and service desk) along the route.

To the best of our knowledge, we are the first to study the route planning queries where the score of a route is computed using not only its total
length but also other relevant attributes such as total price and total waiting time etc. We show that a \queryAbb~query is NP-hard in number of categories
and propose an approximation algorithm to efficiently solve it.  Although it is possible to extend existing outdoor techniques to solve \queryAbb~query in indoor venues, they fail to exploit the properties specific
to indoor venues such as high density of objects in indoor partitions (e.g., thousands of objects in a single store). To address this issue,
we present an efficient algorithm that utilizes a novel dominance-based pruning to significantly reduce the number of possibilities while maintaining
high-quality results. Our extensive experimental study shows the effectiveness of our proposed algorithm. We summarize our contributions below.
\begin{itemize}
\item We propose the \queryName~(\queryAbb) query and show that it is NP-Hard.
\item We present an efficient approximation algorithm to retrieve high-quality results for \queryAbb~queries.
\item We conduct an extensive set of experiments on a real-world shopping center containing real products. The experiments demonstrate
that our algorithm outperforms state-of-the-art technique in terms of running time and quality of results.
Furthermore, our experiments show that the cost of the route generated by our algorithm is at most $20\%$ higher than the cost of the optimal route.
\end{itemize}

The remaining sections are organized as follows,
Section \ref{sec:related} reviews the related work.
In section \ref{sec:problem}, we formulate the problem of \queryAbb~query.
Section \ref{sec:solution} is mainly about the proposed solutions.
Section \ref{sec:experiments} reports the experimental results.
The paper concludes in Section \ref{sec:conclusion}.
\section{Related Work}\label{sec:related}

\subsection{Query processing in indoor space}\label{query_process}
The existing outdoor query processing techniques fall short in indoor space as they do not 
consider unique properties of an indoor space such as hallways and rooms. 
Hence, efficient query processing in indoor space has received a great attention in recent years in which many indexing structures and query processing techniques were proposed. A comprehensive taxonomy for querying indoor data, shortest distance/path, range and k nearest neighbour queries under various settings can be found in \cite{lu2011spatio,xie2015distance,yang2009scalable,yuan2010supporting}. 
RTR-Tree and $TP^2$R-tree \cite{jensen2009indexing} are extensions of R-tree to index trajectories of indoor moving objects. 
Xie~\textit{et al.}~\cite{xie2013efficient} develop a composite indexing structure called $ind$R-tree, that indexes indoor entities into different layers.
D2D graph \cite{yang2010probabilistic} is one of the most notable techniques  which has been used in most of the studies in literature since they enable various query processing techniques in road networks \cite{zhong2015g,lee2012road} to be applied in the indoor space. 
D2D graph represents doors in the indoor space as vertices. A weighted edge between two vertices is created if they are connected to the same indoor partition (e.g., room, hallway) where the edge weight is the indoor distance between the corresponding doors.
Lu~\textit{et al.}~\cite{lu2012foundation} propose a distance aware indoor space data model along with efficient distance computation algorithms. 

Shao~\textit{et al.}~ \cite{shao2016vip} introduce an efficient indexing structure called IP-tree that takes into account unique indoor properties in tree construction and query processing. 
In an IP-tree, adjacent indoor partitions (e.g., rooms, hallways, staircases) are combined to form leaf nodes. Then, the adjacent leaf nodes are combined to form intermediate nodes. This process is iteratively continued until all nodes are combined into a single node (i.e., root node). 
VIP-tree~\cite{shao2016vip} is an improvement of the IP-tree.
Compared to the existing indexing techniques, VIP-tree has demonstrated more efficiency and higher scalability. 
\subsection{Route Planning Queries}\label{route_plan}
A large body of research has been done on developing efficient techniques to process route planning queries. 
Trip planning query (TPQ)~\cite{li2005trip} has source and target locations and a set of categories in which it finds the shortest route starts at the source location, passes through at least one object from each given category and ends at the target location. 
They propose two fast algorithms (a greedy and an integer programming algorithm) based on triangular inequality property of the metric space. 
These solutions take into account only the distance in finding an optimal route while \queryAbb~considers multiple criteria such as static cost. Hence, these solutions cannot be used to process \queryAbb~queries.

Sharifzadeh~\textit{et al.}~\cite{sharifzadeh2008optimal} introduce a variant of TPQ called optimal sequenced route (OSR) query that visits the categories in a particular order given by the user. 
There are several works~\cite{kanza2009route,kanza2010interactive} in literature that study OSR queries.
\queryAbb~is different from OSR since it does not consider a visiting order of the categories. Therefore, these algorithms are not applicable to process \queryAbb~quries.
Cao~\textit{et al.}~\cite{cao2012keyword} introduce another variant called keyword aware optimal route (KOR) search, which covers all user given keywords while satisfying a user specified budget constraint and optimizing objective score of the route. 
Zeng~\textit{et al.}~\cite{zeng2015optimal} find an optimal route such that the keyword coverage is maximized without exceeding a budget constraint. Purpose of such a route is to optimally satisfy the user's weighted preferences. 
Chen~\textit{et al.}~\cite{chen2008multi} study a new type of route planning query called multi-rule partial sequenced route (MRSPSR) query in which users set travelling preferences/restrictions when they issue a query. 
We find these works have different aims compared to \queryAbb~problem.
 
Yao~\textit{et al.}~\cite{yao2011multi} study another variant of route planning query, the multi-approximate-keyword routing (MAKR) query. A MAKR query finds a route with the shortest length such that it covers at least one matching object per given keyword while satisfying string similarity constraints.  
MARK studies a similar problem to \queryAbb~. Thus, we employ an extension of their  approximation solution in our experiments to evaluate our proposed solution.
Shao~\textit{et al.}~\cite{shao2017trip} are the first to study the indoor trip planning queries. They propose an exact solution called VIP-tree neighbour expansion (VNE) that exploits the unique indoor features such as rooms and hallways. Hence, we find that an extension of their solution is inefficient in answering a \queryAbb~query.
\section{Problem Definition} \label{sec:problem}

In this section, we formulate the problem of \queryName~query and prove the hardness of the problem.
Notations used in this paper are summarized in Table \ref{tab:notations}.\\

\begin{definition}[Indoor objects]
Let $p_i \in \mathcal{P}$ be an indoor point representing an indoor object. Each point $p_i$ is associated with a category $c_j \in \mathcal{C}$ and a static score denoted by $s(p_i)$.
\end{definition}

\begin{definition}[Route]
A route $R = \langle p_1,\dots,p_m \rangle$ denotes a path from indoor point $p_1$ to $p_m$ where $\langle p_i,p_{i+1} \rangle$ is the shortest path between two points.  
\end{definition}

\begin{definition}[Travel cost]
Given a route $R = \langle p_1,\dots,p_m \rangle$, the travel cost of route $R$ is computed as follows
\begin{equation}\label{func:travel_cost}
\begin{split}
T(R) = \sum_{i=1}^{m-1} dist(p_i,p_{i+1})
\end{split}
\end{equation}
where $dist(p_i,p_{i+1})$ denotes the indoor distance between two points in route $R$. 
\end{definition}
\begin{definition}[Static cost]
Given a route $= \langle  p_1,\dots,p_m \rangle$, let $R.\psi = \langle c_1,\dots,c_m\rangle$ be the set of categories covered by $R$ where $|R.\psi|= m$ and $p_i$ denotes an indoor point that covers $c_i \in R.\psi$. Hence, the static cost is computed as follows,
\begin{equation}\label{func:static_cost}
\begin{split}
S(R) = \sum_{i=1}^{m} s(p_i)
\end{split}
\end{equation}
where $s(p_i)$ denotes the static score of the indoor point $p_i$.
\end{definition}

\begin{definition}[Cost function]
We determine the cost of a route $R$ in terms of travel cost and static cost, as follows,
 \begin{equation}\label{func:total_cost}
Cost(R) = \alpha \cdot T(R) + (1 - \alpha) \cdot S(R) \enspace 
\end{equation} 
Here, $\alpha$ is a query parameter (user-defined) that lies between 0 and 1 to control the preference of travel cost and static cost.
\end{definition}
\begin{table}[t]
\caption{The summary of notations}
\label{tab:notations}       
\begin{tabular}{|l|l|}
\hline
\textbf{Notation} & \textbf{Definition}  \\
\hline
$p_i$ & An indoor point \\
\hline
$c_j$ & A category  \\
\hline
$s(p_i)$ & The static score of point $p_i$\\
\hline
$q_i$ & A \queryAbb~query \\
\hline
$d_i$ & A door in indoor space \\
\hline
$I_j$ & An indoor partition\\
\hline
$p_s / p_t$ & The start/end point of a route \\
\hline
$\psi$ & A set of categories\\
\hline
$\mathcal{P}_i$ & The set of indoor points of category $c_i$\\
\hline
$p_a \prec_k p_b$ & The point $p_a$ dominates $p_b$ w.r.t door $d_k$\\
\hline
$R_a \prec R_b$ & The route $R_a$ dominates $R_b$\\
\hline
$Dom_a^k$ & The dominated set of point $p_a$ w.r.t door $d_k$\\
\hline
\end{tabular}
\end{table}
\begin{definition}[\queryNameCap~(\queryAbb) query]\label{def:query_form}
Given an indoor space, a \queryName~query $q = \langle p_s, p_t, \psi \rangle$ where $p_s,p_t$ denotes the source point and the target point of the route, and $q.\psi = \langle c_1,\dots,c_m\rangle$ denotes a set of unique categories that describes the user preferences. A route from the point $p_s$ to the point $p_t$, that passes through at least one indoor point from each given category, is called a complete candidate route. Moreover, a \queryAbb~query returns 
a route subject to:
\begin{equation}\label{func:argmin_cost}
 R^{opt}  = \argmin{R\in F(q)} Cost(R)
\end{equation}
where $F(q)$ is the collection of all complete candidate routes for the given query $q$.
\end{definition}
\begin{theorem}
The problem of solving a \queryAbb~query is NP-hard.
\end{theorem}
\begin{proof}
This problem can be reduced from the classical travelling salesman problem (TSP) which is NP-hard. Given a graph in which each edge has a length,  let both start and end points equal to a node $v_0$, each given category is covered by a node $v_i$ with $s_i=0$ where $i=\{1 \dots m\}$ and all the other nodes contain non-query categories. Clearly, the problem of solving \queryAbb~query is identical to the TSP. Thus, the problem of solving \queryAbb~problem is NP-hard.  
\end{proof}
\section{Our Solutions}\label{sec:solution}

\subsection{\algoAbb~Algorithm}\label{sub_sec:gcnn}
A \queryAbb~query can be answered using a brute force approach by conducting an exhaustive search. 
Even though the brute force method guarantees the optimal solution, the exhaustive search is  prohibitively expensive in practice. We devise a novel approximation algorithm called \algoName~(\algoAbb) algorithm to quickly answer a \queryAbb~query. 

\algoAbb~algorithm is a greedy algorithm that greedily adds an indoor point $p$ 
to an existing partial candidate route by minimizing the route cost w.r.t travel and static costs. 
Basically, \algoAbb~algorithm starts from the source point $p_s$ and progressively constructs a candidate route by inserting an indoor point covering one of the uncovered categories. For a given partial candidate route $R = \{p_s,p_{x_1},...,p_{x_j}\}$, the algorithm finds such a point subjected to:
\begin{equation}\label{eq:score}
\begin{aligned}
 score(p_{x_j},p) =  &~\alpha \cdot (dist(p_s,p) + dist(p_{x_j},p) + dist(p,p_t)) \\
 &+ (1 - \alpha) \cdot s(p)
 \end{aligned}
\end{equation}

\begin{equation}\label{eq:cnn}
 cnn(p_{x_j},\mathcal{P}_i) = \argmin{p\in \mathcal{P}_i}  score(p_{x_j},p)
\end{equation}
\begin{equation}\label{eq:pick_point}
 p = \argmin{\forall c_i \in q.\psi\setminus R.\psi} cnn(p_{x_j},\mathcal{P}_i)
\end{equation}
\hfill
\linebreak
where $cnn(p_{x_j},\mathcal{P}_i)$  returns the \textit{category nearest neighbour} point for a given category $c_i$ w.r.t an indoor point $p_{x_j}$. We comprehensively describe the process of obtaining a category nearest neighbour point in Section~\ref{sub_sec:cnn}.
Then, the globally best category nearest neighbour point for the current point $p_{x_j}$ is determined using Equation~\eqref{eq:pick_point} and $R$ is updated to $R = \{p_s,p_{x_1},...,p_{x_j},p\}$. The algorithm terminates when $R$ turns into a complete route where all the query categories are covered.
In order to determine such an optimal route, we can maintain a min-priority queue where a partial candidate route $R$ is enqueued into the queue by determining the key value as follows: $Cost(R) + dist(p_s,p) + dist(p,p_t)$ where $p$ is the recently inserted point.
Whenever a candidate route is dequeued from the queue, we find category nearest neighbour points for each uncovered category and generate new candidate routes. Then, the set of new candidate routes are enqueued into the queue. Intuitively, the candidate route which is dequeued first in next iteration is the answer to Equation~\eqref{eq:pick_point}.

\begin{figure}[h]
	\centering
	\includegraphics[width=0.4\textwidth]{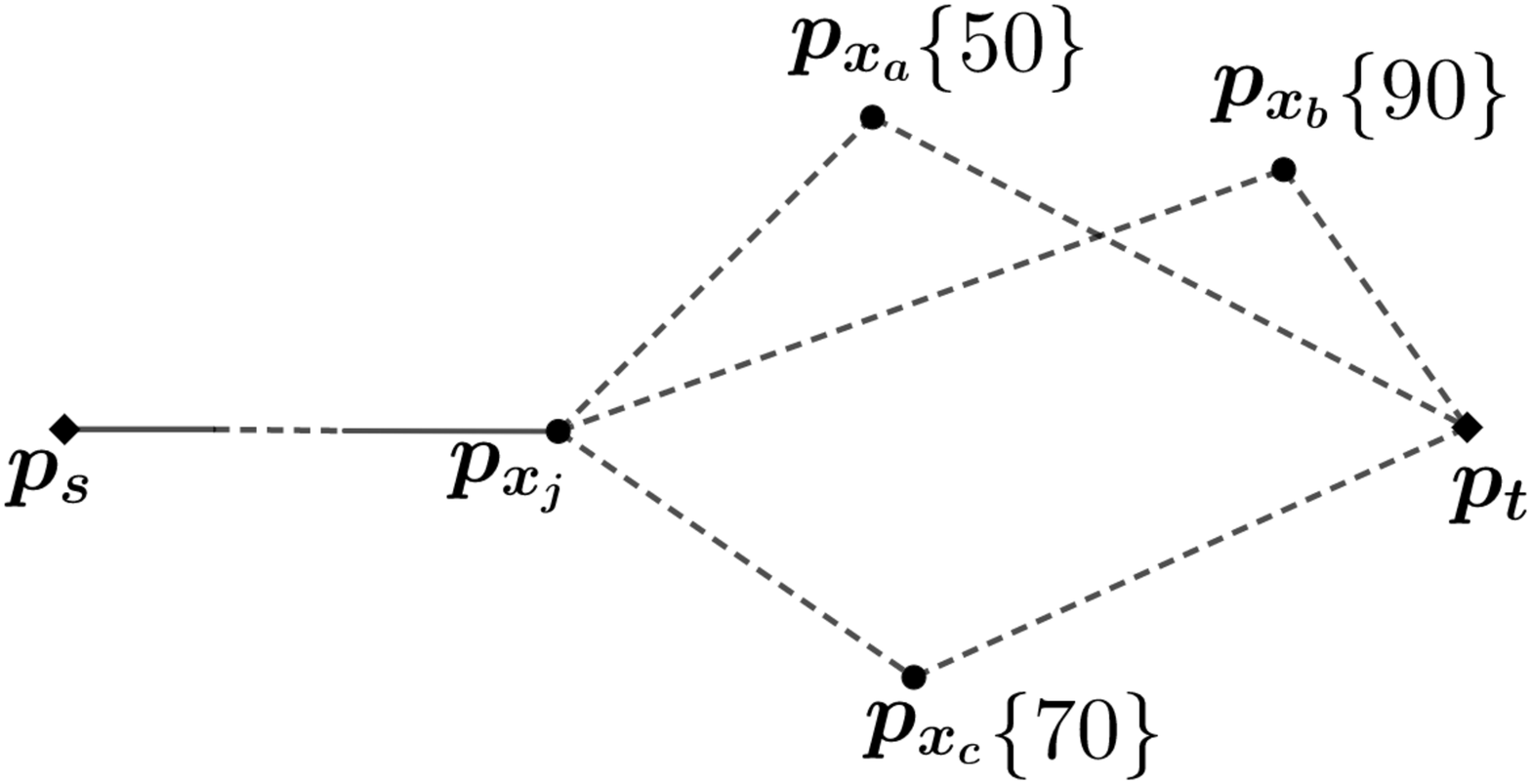}
	\caption{Example of point selection }
	\label{fig:gcnn_algo}
\end{figure}

As Algorithm~\ref{algo:gcnn_algo} illustrates, initially, we enqueue a route $R = \{p_s \}$ with zero as the key value. 
We terminate the algorithm either when the queue is empty (line 4) or an optimal route is found (line 8-10). In each iteration, we dequeue a candidate route $R^* = \{p_s,...,p_{x_j}\}$ from the queue (line 5) which essentially provides the answer to Equation~\eqref{eq:pick_point} of the previous iteration. 
After a candidate route is dequeued, we clear the min-priority queue by dequeuing all the routes (line 6). This allows us to maintain the current optimal partial candidate route in each iteration. Next, the set of uncovered categories, i.e., $\Psi$, is obtained (line 7). 
Then for each uncovered category, we get the category nearest neighbour point $p$ using Equation~\eqref{eq:cnn} and generate a new candidate route by inserting that point into the current candidate route (line 11-13). The key value of a route is determined by taking into account both route cost and distances between point $p$ and start/end points, i.e., $dist(p_s,p)$ and $dist(p, p_t)$, (line 14).
Each route is then enqueued into the queue with its key value (line 15). Finally, the optimal route for the given \queryAbb~query is returned (line 16).
\begin{algorithm}[t]
	\begin{small}
	\KwData{A \queryAbb~query $q = \{p_s, p_t, \psi\}$}
	\KwResult{An optimal route $R$}
	
	$\mathcal{Q} \gets \emptyset$;\\
	$R \gets \{p_s \}$;\\
	$\mathcal{Q}.\textit{enqueue}(R,0)$;\\
	\While{$\mathcal{Q}$ is \textbf{NOT} empty}{
		
		$R^* \gets \mathcal{Q}.\textit{dequeue}()$\tcp*{ Equation~\eqref{eq:pick_point}}
		\tcp{Let $R^* = \{p_s,...,p_{x_j}\}$}
		$\mathcal{Q}.\textit{clear}()$;\\
		$\Psi \gets q.\psi \setminus R.\psi$;\\
		\If{$\Psi = \emptyset$}{\tcp{when route cover all categories}
			$R \gets \{p_s,...,p_{x_j}, p_t\}$;\\
			\Break;
		}

		\ForEach{category $c_i \in \Psi$}{
			$p \gets cnn(p_{x_j},\mathcal{P}_i)$\tcp*{ Equation~\eqref{eq:cnn}}
			$R^*_i \gets \{p_s,...,p_{x_j}, p\}$;\\
			$key \gets Cost(R^*_i) + dist(p_s,p) + dist(p, p_t)$;\\
			$\mathcal{Q}.\textit{enqueue}(R^*_i,key)$;\\
		}
			
	}			
	\Return $R$;\\					

	\end{small}
	\caption{\algoAbb~Algorithm}
\label{algo:gcnn_algo}
\end{algorithm}
\setlength{\textfloatsep}{10pt}
For example, Figure~\ref{fig:gcnn_algo} shows a route $R = \{p_s,...,p_{x_j}\}$ where $R.\psi = \{c_3\}$. Let $p_{x_a}, p_{x_b}$ and $p_{x_c}$ be indoor points  where $p_{x_a}, p_{x_c}$ belong to category $c_1$, i.e., $p_{x_a}, p_{x_c} \in \mathcal{P}_1$, and $p_{x_b}$ belongs to category $c_2$, i.e., $p_{x_b} \in \mathcal{P}_2$. The score of each point w.r.t Equation~\eqref{eq:score} is mentioned next to the point.
Assume that $q.\psi = \{c_1, c_2, c_3\}$ and $R$ be the recently dequeued candidate route. Then 
\algoAbb~algorithm finds the category nearest neighbour point for each uncovered category, i.e., $c_1,c_2$, using Equation~\eqref{eq:cnn}. 
Hence, the points $p_{x_a}$ and $p_{x_b}$ are selected as $cnn(p_{x_j},\mathcal{P}_1)$ and $cnn(p_{x_j},\mathcal{P}_2)$ respectively. 
Then, new candidate routes $R^*_1 = \{p_s,...,p_{x_j}, p_{x_a}\} $ and $R^*_2 = \{p_s,...,p_{x_j}, p_{x_b}\} $ are generated accordingly and enqueued into the queue. 
In the next iteration, $R^*_1$ is dequeued first satisfying Equation~\eqref{eq:pick_point}. 
\subsubsection{Category Nearest Neighbour}\label{sub_sec:cnn}
The category nearest neighbour of point $p$ is the closest point to $p$ w.r.t both travel and static costs. 
In order to obtain the category nearest neighbour point covering category $c_i$ for a given point $p_{x_j}$, i.e., $cnn(p_{x_j},\mathcal{P}_i)$, every indoor point $p$ belongs to the particular category $c_i$, i.e., $p \in \mathcal{P}_i$, is ranked using Equation~\eqref{eq:score}. As Equation~\eqref{eq:cnn} depicts, the point with the minimum ranking score is selected eventually. 

Determining an optimal route in \algoAbb~algorithm is very expensive since all the related indoor points belong to uncovered categories are ranked in each iteration to find category nearest neighbours. 
Thus, the number of times, the ranking operation is executed in obtaining an optimal route is $\mathcal{O}(n \cdot m^2)$, where $n$ is the average number of indoor points per category and $m$ is the total number of query categories.
Clearly, the performance of \algoAbb~algorithm decreases as $m$ and $n$ are increased. Since, $m$ is a query parameter, the performance of the algorithm
can be accelerated by reducing the number of related indoor points (i.e., $n$) in the indoor space. Thus, we introduce a novel pruning technique that eliminates the indoor points which are highly unlikely to be selected in determining an optimal route.
We use an extension of VIP-tree~\cite{shao2016vip} called \textit{inverted VIP-tree} as our indexing structure. In order to support category-based filtering, we modified VIP-tree by implementing an inverted file at each tree node. For example, an inverted file in a leaf node consists of a list of all the unique categories that appear in any indoor partition, and for each category, a list of indoor partitions in which it appears. 
Moreover, we maintain an additional list at each tree node which consists of the minimum static score of each category appear in the tree node. This enables simultaneous travel cost and static cost based filtering. Thus, a category nearest neighbour point is retrieved in an efficient manner. 
\subsection{Dominance-based Pruning}\label{sub_sec:dom_based_prune}
As we discussed in previous section, the performance of \algoAbb~algorithm can be accelerated by reducing the number of indoor points visited by the algorithm in query processing. Thus, we introduce a novel pruning technique called dominance-based pruning that eliminates objects that are highly unlikely to be selected in constructing the optimal solution. The dominance-based pruning technique utilizes the unique properties of an indoor space such as partitions in which it identifies the incompetent points in each indoor partition and prune them accordingly. Before we present the pruning technique, we introduce following definitions. 
\begin{definition}[Point Dominance]\label{def:point_dom}
Let $p_a$ and $p_b$ be points belong to category $c_i$, i.e., $p_a,p_b \in \mathcal{P}_i$, reside in an indoor partition $I$. Let $d_s$ be one of the doors of $I$. Then, the indoor point $p_a$ dominates $p_b$ w.r.t door $d_s$, denoted by $p_a \prec_{d_s} p_b$, \textit{if and only if} $dist(d_s,p_a) < dist(d_s,p_b)$ and $s(p_a) < s(p_b)$. 
\end{definition}
\begin{definition}[Dominated Set]\label{def:dominated_set}
Let $p_a$ be a point belongs category $c_i$, i.e., $p_a \in \mathcal{P}_i$. 
Then dominated set of the point $p_a$ w.r.t door $k$ is defined as follows,
\begin{equation}\label{eq:dom_set}
Dom_a^k = \bigcup_{p_a \prec_{d_k} p_j,\forall p_j \in \mathcal{P}_i} p_j
\end{equation}
\end{definition}
The dominance of a point over another point can be decided only if both points belong to the same category and reside in the same indoor partition. 
As Definition~\ref{def:point_dom} depicts, for a given door $d_s$ and two indoor points $p_a, p_b \in \mathcal{P}_i$, if the point $p_a$ is closer to the door $d_s$ than the point $p_b$ and also has a static cost less than $p_b$,
then  $p_a$ dominates $p_b$ w.r.t the door $d_s$. Moreover, according to Definition~\ref{def:dominated_set},  the point $p_b$ belongs to the dominated set of $p_a$ w.r.t door $d_k$.

\begin{definition}[Route Dominance]\label{def:path_dom}
Let $R_a$ and $R_b$ be routes inside an indoor partition $I$, start from door $d_s$ and end at door $d_t$ where $R_a.\psi = R_b.\psi$. Then, $R_a$ dominates $R_b$ (denoted by $R_a \prec R_b$) \textit{if and only if} $Cost(R_a) < Cost(R_b)$.
\end{definition}
A route can dominate another route only if both routes are inside the same partition, the starting and ending doors are same, and covering the same set of categories.  
According to Definition~\ref{def:path_dom}, if the cost of route $R_a$ is less than the cost of route $R_b$, then route $R_a$ dominates $R_b$.
Next, we present four important theorems that help to derive our pruning rules. 
Note that, for all these theorems and pruning rules, we assume that $q.\psi = \{c_m,c_n \}$ and an indoor partition $I$ consist of two doors $d_s,d_t$. Also, when we say $\mathcal{P}_i$, it means set of points of the indoor partition $I$ that belongs to category $c_i$.

\begin{theorem}\label{thm:theorem_1}
Let routes $R_a = \langle d_s, p_a, p_x, d_t \rangle$ and $R_b = \langle d_s, p_b, p_x, d_t \rangle$ where $p_a,p_b \in \mathcal{P}_m$ and $p_x \in \mathcal{P}_n$. Then, $R_a \prec R_b$ only if $p_a \prec_{d_s} p_b$ and $dist(p_a,p_x) < dist(p_b,p_x)$. 
\end{theorem}
\begin{proof}
For the given routes $R_a$ and $R_b$, if $p_a$ dominates $p_b$ then $dist(d_s,p_a)+s(p_a) < dist(d_s,p_b)+s(p_b)$. Also, we know that, $dist(p_a,p_x) < dist(p_b,p_x)$. By adding both inequalities,  $dist(d_s,p_a) + s(p_a) + dist(p_a,p_x) < dist(d_s,p_b) + s(p_b) + dist(p_b,p_x)$. Furthermore, $dist(d_s,p_a)+dist(p_a,p_x)+s(p_a)+dist(p_x,d_t)+s(p_x) < dist(d_s,p_b)+s(p_b)+dist(p_b,p_x)+dist(p_x,d_t)+s(p_x)$. And, $Cost(R_a) < Cost(R_b)$. Hence, $R_a \prec R_b$. 
\end{proof}

\begin{figure}
	\subfloat[]{\includegraphics[width=0.25\textwidth]{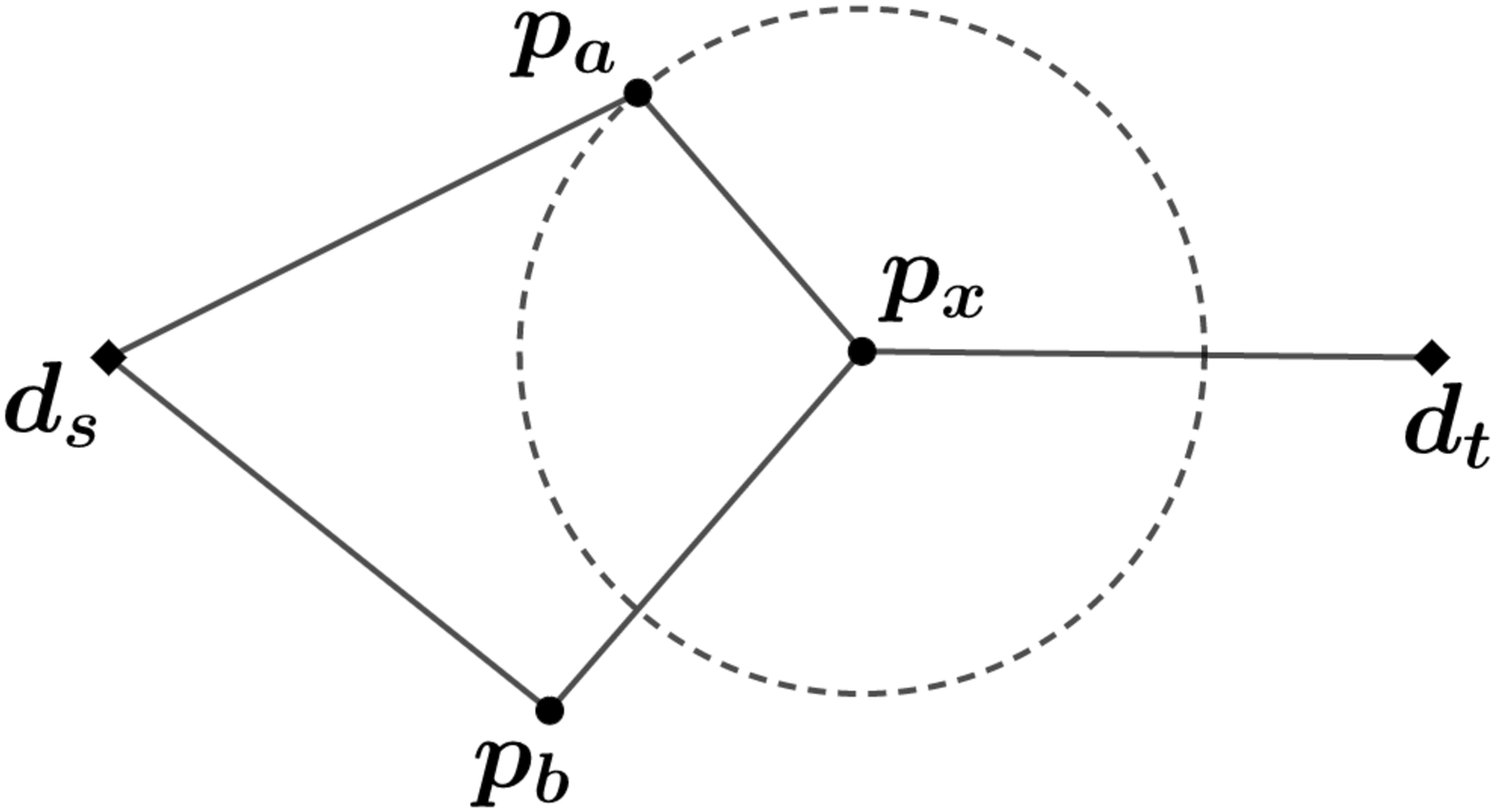}}
	\subfloat[]{\includegraphics[width=0.25\textwidth]{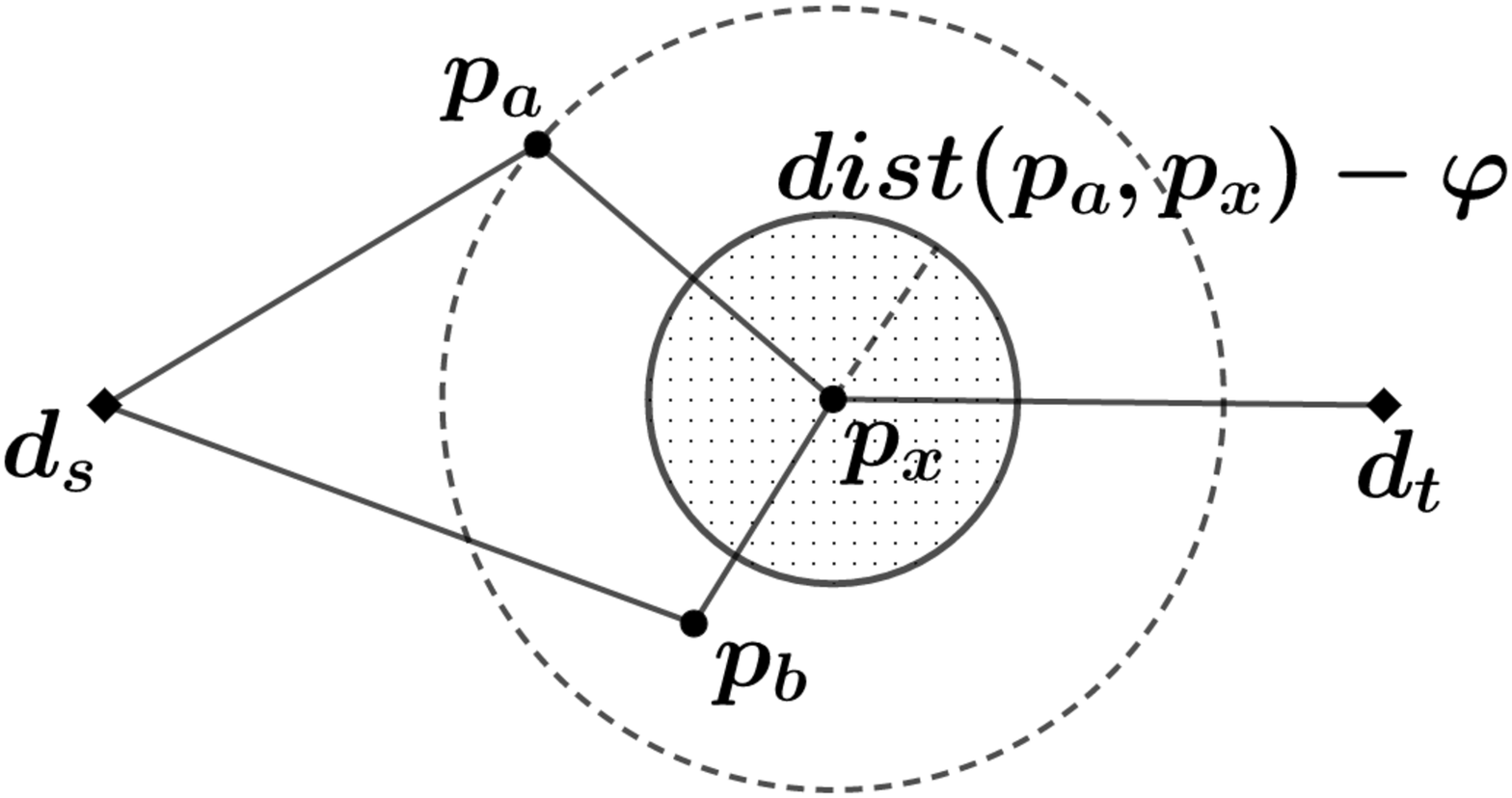}}   
	
	\caption{Example of Theorem~\ref{thm:theorem_1} and Theorem~\ref{thm:theorem_2}}
	\label{fig:theorem_1_2}       
\end{figure}

\begin{theorem}\label{thm:theorem_2}
Let routes $R_a = \langle d_s, p_a, p_x, d_t \rangle$ and $R_b = \langle d_s, p_b, p_x, d_t \rangle$ where $p_a,p_b \in \mathcal{P}_m$ and $p_x \in \mathcal{P}_n$, and $dist(p_a,p_x) \geq dist(p_b,p_x)$. Then, $R_a \prec R_b$ only if $p_a \prec_{d_s} p_b$ and $dist(p_a,p_x) - \varphi < dist(p_b,p_x)$ where $\varphi = dist(d_s,p_b)+s(p_b)- dist(d_s,p_a)-s(p_a)$.
\end{theorem}
\begin{proof}
We prove this by contradiction, Assume $R_b \prec R_a$ and  $p_b \prec_{d_s} p_a$. 
Then, $dist(d_s,p_a)+s(p_a) > dist(d_s,p_b)+s(p_b)$. Hence, $0> \varphi$. Also, we know $dist(p_a,p_x) \geq dist(p_b,p_x)$. By above two inequalities, $dist(p_a,p_x) - \varphi > dist(p_b,p_x)$. Therefore, it must be the case that our assumption is false. So $R_a \prec R_b$ when $p_a \prec_{d_s} p_b$ and $dist(p_a,p_x) - \varphi < dist(p_b,p_x)$.
\end{proof}
For given $q.\psi = \{c_m,c_n \}$, the dominance of route $R_a = \langle d_s, p_a, p_x, d_t \rangle$ over $R_b = \langle d_s, p_b, p_x, d_t \rangle$ can be guaranteed if the point $p_a$ dominates $p_b$ and $p_a$ is closer to $p_x$ than $p_b$ (See Figure~\ref{fig:theorem_1_2}(a)).
Theorem~\ref{thm:theorem_2} takes into account an instance where the point $p_b$ is closer to $p_x$ than $p_a$. In this case, $R_a \prec R_b$ can be guaranteed only if the point $p_b$ resides outside the distance threshold $dist(p_a,p_x) - \varphi$ as Figure~\ref{fig:theorem_1_2}(b) illustrates.\\
\hfill
\linebreak
\textbf{For multiple objects.} Assume that there is another point $p_j \in \mathcal{P}_m$ within the distance $dist(p_a,p_x)$, where $p_b \prec_{d_s} p_j$. 
If $dist(p_a,p_x)- \varphi < dist(p_j,p_x)$ where $\varphi = dist(d_s,p_b)+s(p_b)- dist(d_s,p_a)-s(p_a)$, then point $p_j$ can be ignored since a route via $p_j$ does not dominate $R_a$. If the indoor points, i.e., $p_b, p_j$, are visited based on \textit{dominance order}, then distance threshold, i.e., $dist(p_a,p_x) - \varphi$, is guaranteed to be an upper bound as $\varphi$ is always a lower bound. Visiting the indoor points based on dominance order means that always a point $p$ is visited before visiting a point dominated by $p$. 
Moreover, if $dist(p_a,p_x)- \varphi > dist(p_j,p_x)$, then $\varphi$ needs to be updated w.r.t point $p_j$ and checked for $dist(p_a,p_x)- \varphi < dist(p_j,p_x)$. Similarly, all points need to be verified if there is more. Then we can guarantee that $R_a$ dominates $R_j$ where $R_j = \langle d_s, p_j, p_x, d_t \rangle$, $\forall p_j \in \mathcal{P}_m$. 

\begin{theorem}\label{thm:theorem_3}
Let routes $R_a = \langle d_s, p_a, p_x, d_t \rangle$ and $R_b = \langle d_s, p_b, p_y, d_t \rangle$ where $p_a, p_b \in \mathcal{P}_m$ and $p_x, p_y \in \mathcal{P}_n$. Then, $R_a \prec R_b$ only if $p_a \prec_{d_s} p_b$, $p_x \prec_{d_t} p_y$ and $dist(p_a,p_x) < dist(p_b,p_y)$. 
\end{theorem}
\begin{proof}
For the given routes $R_a$ and $R_b$, if $p_a$ dominates $p_b$ and $p_x$ dominates $p_y$, then $dist(d_s,p_a)+s(p_a) < dist(d_s,p_b)+s(p_b)$ and $dist(d_t,p_x)+s(p_x) < dist(d_t,p_y)+s(p_y)$ respectively. Also, we know that $dist(p_a,p_x) < dist(p_b,p_x)$. By adding them, $dist(d_s,p_a)+s(p_a)+dist(p_a,p_x)+dist(d_t,p_x)+s(p_x) < dist(d_s,p_b)+s(p_b)+dist(p_b,p_y)+dist(d_t,p_y)+s(p_y)$. And, $Cost(R_a) < Cost(R_b)$. Hence, $R_a \prec R_b$. 
\end{proof}

\begin{theorem}\label{thm:theorem_4}
Let routes $R_a = \langle d_s, p_a, p_x, d_t \rangle$ and $R_b = \langle d_s, p_b, p_y, d_t \rangle$ where $p_a, p_b \in \mathcal{P}_m$ and $p_x, p_y \in \mathcal{P}_n$, and $dist(p_a,p_x) \geq dist(p_b,p_y)$. Then, $R_a \prec R_b$ only if $p_a \prec_{d_s} p_b$, $p_x \prec_{d_t} p_y$ 
and $dist(p_a,p_x) - \hat{\varphi} < dist(p_b,p_x)$ where $\hat{\varphi} = dist(d_s,p_b)+s(p_b)+dist(p_y,d_t)+s(p_y) - dist(d_s,p_a)-s(p_a)-dist(p_x,d_t)-s(p_x) $.

\end{theorem}
\begin{proof}
We prove this by contradiction, Assume $R_b \prec R_a$, $p_b \prec_{d_s} p_a$ and  $p_y \prec_{d_s} p_x$. 
Then, $dist(d_s,p_a)+s(p_a) > dist(d_s,p_b)+s(p_b)$ and $dist(d_t,p_x)+s(p_x) > dist(d_t,p_y)+s(p_y)$. Hence, $0> \varphi$. 
Also, we know $dist(p_a,p_x) \geq dist(p_b,p_y)$. By above two inequalities, $dist(p_a,p_x) - \varphi > dist(p_b,p_y)$. Therefore, it must be the case that our assumption is false. So $R_a \prec R_b$ when $p_a \prec_{d_s} p_b$, $p_x \prec_{d_t} p_y$ 
and $dist(p_a,p_x) - \hat{\varphi} < dist(p_b,p_x)$.
%
\end{proof}
\begin{figure}
	\subfloat[]{\includegraphics[width=0.25\textwidth]{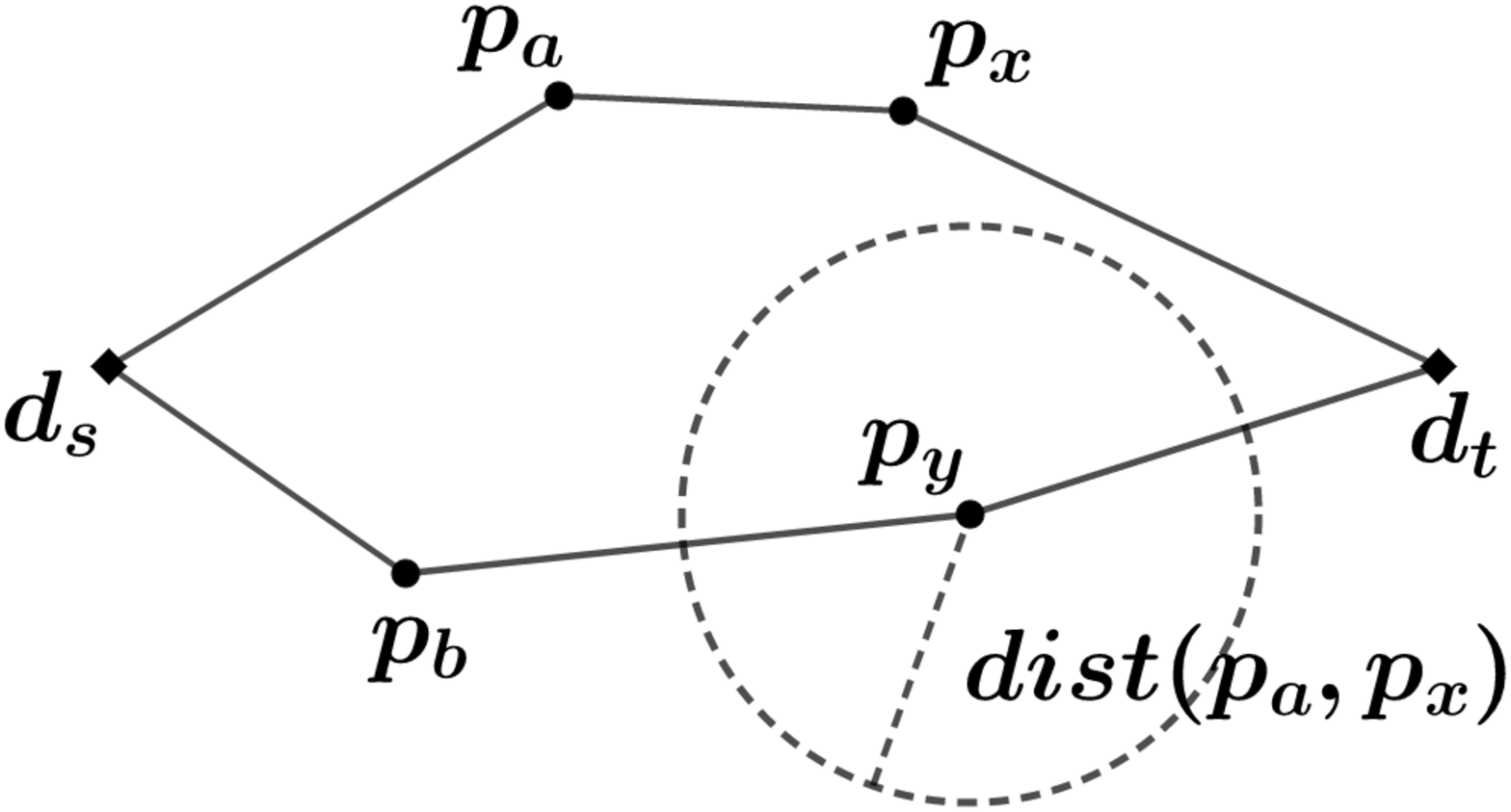}}
	\subfloat[]{\includegraphics[width=0.25\textwidth]{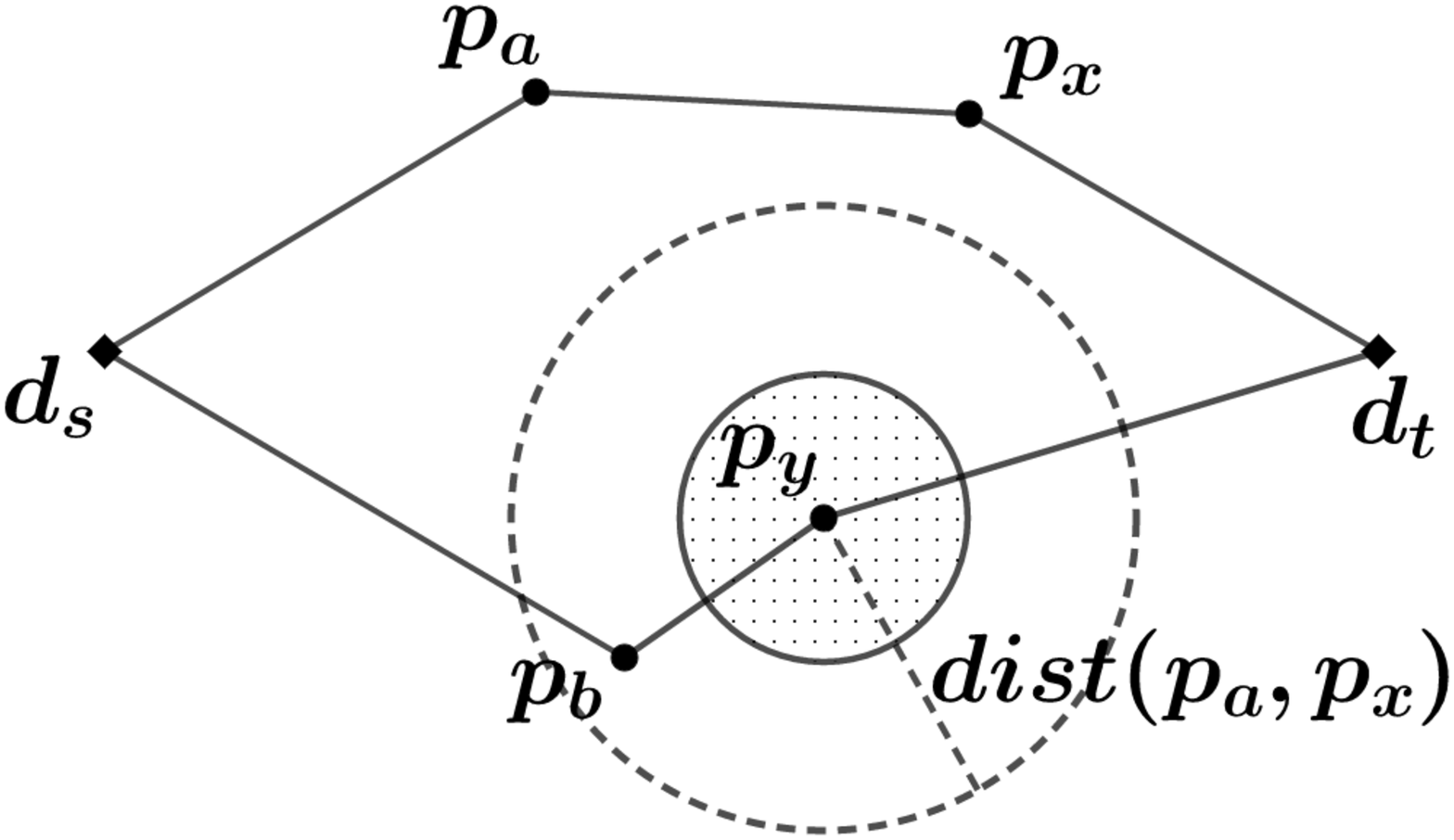}}   
	
	\caption{Example of Theorem~\ref{thm:theorem_3} and Theorem~\ref{thm:theorem_4}}
	\label{fig:theorem_3_4}       
\end{figure}
As Figure~\ref{fig:theorem_3_4}(a) shows, Theorem~\ref{thm:theorem_3} guarantees that a route via points $p_b \in Dom_a^s$ and $p_y \in Dom_x^t$ , i.e., $R_b = \langle d_s, p_b, p_y, d_t \rangle$, cannot dominate a route via corresponding points $p_a$ and $p_x$, i.e., $R_a = \langle d_s, p_a, p_x, d_t \rangle$ when the distance between points $p_a$ and $p_x$ is less than the distance between points in corresponding dominated sets. Theorem~\ref{thm:theorem_4} explains an instance (See See Figure~\ref{fig:theorem_3_4}(b)) where $dist(p_a, p_x) \geq dist(p_b, p_y)$. In this case $R_a$ dominates $R_b$ if the distance between the points in dominated sets is greater than the particular distance threshold, i.e.,  $dist(p_a, p_x) - \hat{\varphi}$.
 
Next, we proceed to introduce our pruning rules which are derived from the aforementioned theorems. These pruning rules help to filter all the points in an indoor partition that are highly unlikely to be selected in generating an optimal route.
 
\begin{prune}\label{prune:rule_1}
Let $p_i \in Dom_a^s$ and $p_b = nn(p_a)$ where $p_a \in \mathcal{P}_m$ and $p_b \in \mathcal{P}_n$. Then, the points $p_a, p_b$ are selected and a point $p_j \in Dom_b^t$ is pruned only if $p_a = nn(p_b)$ and $dist(p_a,p_b) < dist(p_i,p_j)$. 
\end{prune} 
\begin{proof}
According to Theorem~\ref{thm:theorem_1}, when $p_i \in Dom_a^s$ and $p_b = nn(p_a)$, then $Cost(R_x)<Cost(R_y)$ where  $R_x = \langle d_s, p_a, p_b, d_t \rangle$ and $R_y = \langle d_s, p_i, p_b, d_t \rangle$. Also, when  $p_a = nn(p_b)$ and $dist(p_a,p_b) < dist(p_i,p_j)$, 
Theorem~\ref{thm:theorem_3}~guarantees $Cost(R_x)<Cost(R_z)$ where $R_z = \langle d_s, p_i, p_j, d_t \rangle$. Thus, the point $p_j \in Dom_b$ can be pruned since any path covering category $c_m$ and $c_n$ via $p_j$ is dominated by the path $R_x$. 
\end{proof}
\begin{prune}\label{prune:rule_2}
Let $p_i \in Dom_a^s$ and $p_b = nn(p_a)$ where $p_a \in \mathcal{P}_m$ and $p_b \in \mathcal{P}_n$. Then,  the points $p_a, p_b$ are selected and a point $p_j \in Dom_b^t$ is pruned when $p_a \neq nn(p_b)$ and $dist(p_a,p_b) < dist(p_i,p_j)$ only if $Cost(R_x)<Cost(R_y)$ where  $R_x = \langle d_s, p_a, p_b, d_t \rangle$ and $R_y = \langle d_s, p_i, p_b, d_t \rangle$. 
\end{prune}
\begin{prune}\label{prune:rule_3}
Let $p_i \in Dom_a^s$ and $p_b = nn(p_a)$ where $p_a \in \mathcal{P}_m$ and $p_b \in \mathcal{P}_n$. Then,  the points $p_a, p_b$ are selected and a point $p_j \in Dom_b^t$ is pruned when $p_a = nn(p_b)$ and $dist(p_a,p_b) \geq dist(p_i,p_j)$ only if $cost(R_x)<cost(R_y)$ where  $R_x = \langle d_s, p_a, p_b, d_t \rangle$ and $R_y = \langle d_s, p_i, p_j, d_t \rangle$. 
\end{prune} 
\begin{prune}\label{prune:rule_4}
Let $p_i \in Dom_a^s$ and $p_b = nn(p_a)$ where $p_a \in \mathcal{P}_m$ and $p_b \in \mathcal{P}_n$. Then,  the points $p_a, p_b$ are selected and a point $p_j \in Dom_b^t$ is pruned when $p_a \neq nn(p_b)$ and $dist(p_a,p_b) \geq dist(p_i,p_j)$ only if $Cost(R_x)<Cost(R_y)$ where  $R_x = \langle d_s, p_a, p_b, d_t \rangle$ and $R_y = \langle d_s, p_i, p_k, d_t \rangle$ for given $p_k \in Dom_b \cup p_b$. 
\end{prune} 
The Pruning Rule~\ref{prune:rule_2},~\ref{prune:rule_3} and~\ref{prune:rule_4} can also be easily proven using the aforementioned theorems. We have omitted the proofs due to the space limitation.
\begin{algorithm}[t]
	\begin{small}
	\KwData{ Doors $d_s, d_t$, A pair of categories $c_a,c_b$}
	\KwResult{Sets of points $\mathcal{S}_a, \mathcal{S}_b$}

	\While{$\mathcal{P}_a \neq \emptyset$ OR $\mathcal{P}_b \neq \emptyset$}{ 
							
		$p_i \gets \textit{getPoint }(d_s,\mathcal{P}_a)$\tcp*{Dominance order}
		$\mathcal{S}_a \gets \mathcal{S}_a \cup p_i$\tcp*{$p_i$ is selected}
		$\mathcal{P}_a \gets \mathcal{P}_a \setminus \mathcal{S}_a$;\\
		$\overline{\mathcal{P}_b} \gets \mathcal{P}_b$;\\
		\While{$\overline{\mathcal{P}_b} \neq \emptyset$}{
			
			$p_j \gets \textit{NextNN }(p_i)$ where $p_j \in \overline{\mathcal{P}_b}$;\\		
			$U_j \gets \forall p_k \in P_a \setminus S_a$ where $dist(p_i,p_j)>dist(p_k,p_j)$;\\
		
			\If{$U_j = \emptyset$}{
				$\mathcal{S}_b \gets \mathcal{S}_b \cup p_j$\tcp*{$p_j$ is selected}
				$Dom_j^t \gets $ set of points dominated by $p_j$;\\
				$\overline{\mathcal{P}_b} \gets \overline{\mathcal{P}_b} \setminus (Dom_j^t \cup p_j)$;\\
				$\mathcal{P}_b \gets \mathcal{P}_b \setminus \textit{prunePonits }(p_i, p_j, \mathcal{P}_a, Dom_j^t)$
			}
			\Else{
				\ForEach{$p_k \in U_j$}{\tcp{Ascending order}
					
					$\varphi = dist(d_s,p_k)+s(p_k)- dist(d_s,p_i)-s(p_i)$;\\
					$V_k \gets \forall p_m \in  U_j$ where $dist(p_m,p_j)< dist(p_i,p_j) - \varphi$;\\					
					\If{$V_k = \emptyset$}
					{	
						$\mathcal{S}_b \gets S_b \cup p_j$\tcp*{$p_j$ is selected}
						$Dom_j^t \gets $ set of points dominated by $p_j$;\\
						$\overline{P_b} \gets \overline{\mathcal{P}_b}- Dom_j^t$;\\
						$\mathcal{P}_b \gets \mathcal{P}_b \setminus \textit{prunePonits }(p_i, p_j, \mathcal{P}_a, Dom_j^t)$
					}
					\Else{
						\If{ $p_k \in V_k$}
						{
							\Break \tcp*{$p_j$ is not selected}
						}
						\Else{
							$U_j \gets U_j \setminus V_k$ \tcp*{remove points outside the range }
							
						}
					}					
				}							
			}			
		}
	}

	\Return $\mathcal{S}_a, \mathcal{S}_b$
	\end{small}
	\vspace{2mm}
	\caption{$\textit{selectPoints }(\dots)$}
\label{algo:select_points}
\end{algorithm}

\begin{definition}[Dominant point]\label{def:dominant_point}
An indoor point $p$ is called a dominant point if it is highly likely to be selected in generating an optimal route. 
\end{definition}


Simply, a dominant point is a point that is selected by a pruning rule while a non-dominant point is a point which is never selected by a pruning rule.  
Accordingly, the pruning rules are capable of identifying the dominant points while pruning the non-dominant points as they incapable of generating better routes than the routes of dominant points. i.e., dominant routes. 
Due to the space limitations, we provide an example only for Pruning Rule~\ref{prune:rule_1}.
Let $I$ be an indoor partition consist of two doors $d_s, d_t$ and three indoor points $p_a \in \mathcal{P}_m$ and $p_b, p_c \in \mathcal{P}_n$ where $p_b \prec_{d_t} p_c$. Assume that a user who wants to find a route from $d_s$ to $d_t$ covering $c_m, c_n$ categories, visit the point $p_a$ first. Then either the point $p_b$ or $p_c$ needs to be visited before visiting door $d_t$ to get a complete route. 
Assume that the user visits the point $p_b$. Then, according to Pruning Rule~\ref{prune:rule_1}, the point $p_c$ can be pruned only if $dist(p_a, p_b) < dist(p_a, p_c)$. Because, the route $R_x = \langle d_s, p_a, p_b, d_t \rangle$ dominates $R_y = \langle d_s, p_a, p_c, d_t \rangle$. Moreover, the points $p_a, p_b$ are selected as dominant points. 
Otherwise, when $p_c$ is closer to $p_a$ than $p_b$ and $dist(p_a, p_c)$ is less than the threshold distance, i.e., $dist(p_a, p_b) - \varphi$, then the points $p_a, p_c$ is selected and the point $p_b$ is pruned.

\begin{algorithm}[t]
	\begin{small}
	\KwData{Points $p_i, p_j$, Sets of points $\mathcal{P}_a, Dom_j^k \subseteq \mathcal{P}_b$}
	\KwResult{A set of points $ S_b $}

	\ForEach{$p_k \in Dom_j^k$}{\tcp{According to the dominance order}
			
		$p_m \gets NN(p_k)$ where $p_m \in \mathcal{P}_a$;\\
		\eIf{$dist(p_i,p_j) < dist(p_k,p_m)$}{
			
			$S_b \gets S_b \cup p_k$ \tcp*{Theorem~\ref{thm:theorem_3}}
		}
		{
			$\hat{\varphi} = dist(d_s,p_m)+s(p_m)+dist(p_k,d_t)+s(p_k) - dist(d_s,p_i)-s(p_i)-dist(p_j,d_t)-s(p_j) $;\\
			\If{$d(p_k,p_m)> dist(p_i,p_j) - \hat{\varphi}$}{
				$S_b \gets S_b \cup p_k$ \tcp*{Theorem~\ref{thm:theorem_4}}
			}
		}
		
	}

	\Return $ S_b $
	\end{small}
	\caption{$\textit{prunePoints }(\dots)$}
\label{algo:prune_points}
\end{algorithm}
Now, we proceed to explain Algorithm~\ref{algo:select_points} which utilizes aforementioned pruning rules to select dominant points of a given indoor partition w.r.t. a given pair of categories and pair of doors. 
Clearly, the visiting order of the points is crucial in applying the aforementioned pruning rules. Thus, we visit points based on the dominance order, i.e., a point $p_a$ is visited before visiting a point $\hat{p_a} \in Dom_a^k$, by utilizing $\textit{getPoint }(d_i,\mathcal{P}_n)$ which 
returns a point $p \in \mathcal{P}_n$ with the minimum score w.r.t. a monotonic ranking function $f(p) = dist(d_i,p) + s(p)$. 
Initially, point sets  $\mathcal{P}_a$ and $\mathcal{P}_b$ contains the indoor points belong to category $c_a$ and $c_b$ respectively. If a point is either selected or pruned, then the point is removed from the corresponding point set. Hence, 
the algorithm is terminated when one of the point sets, i.e., $\mathcal{P}_a$ or $\mathcal{P}_b$, is empty (line 1).  
First, we obtain a point $p_i$ belong to category $c_a$ by utilizing $\textit{getPoint }(\cdot)$ (line 2). Then, the point $p_i$ is selected as a dominant point of category $c_a$.
We maintain a temporary point set $\overline{\mathcal{P}_b}$ to maintain the non-pruned set of points per iteration. The inner while loop terminates when $\overline{\mathcal{P}_b}$ is empty indicating that all the dominant points belong to category $c_b$ based on $p_i$ is selected while the rest of the points is pruned (line 6). After point $p_i$ is selected we find the closest point to $p_i$, i.e., the point $p_j$, that belongs to category $c_b$. Then, we check whether there are any points closer to $p_j$ than $p_i$ that belong to category $c_a$. If not we select $p_j$ and prune all the points dominated by $p_j$ according to the Pruning Rule~\ref{prune:rule_1} and Pruning Rule~\ref{prune:rule_3} (line 9-13). Else, each point that is closer to $p_j$ than $p_i$ is verified and pruned according to the Pruning Rule~\ref{prune:rule_2} and Pruning Rule~\ref{prune:rule_4} (line 15-27). Note that, each time we update the $\overline{\mathcal{P}_b}$ by removing the dominated points. Because, if a point is dominated then they cannot be selected in the same iteration. But, we update the $\mathcal{P}_b$ set after verifying that a point can be pruned for good by utilizing Algorithm~\ref{algo:prune_points}. 
While we iterate the set of closer points, if one of the points is within the distance threshold, then the point $p_j$ is not selected as another point in category $c_a$ creates a better route with $p_j$ (line 24-25). Also, we can remove all the points outside the current distance threshold since it provides an upper bound as we explained earlier (line 27).
Finally, it returns the selected set of dominant points per given category (line 28).  
 
Algorithm~\ref{algo:prune_points} identifies the indoor points that can be pruned based on Theorem~\ref{thm:theorem_3} and~\ref{thm:theorem_4}. For a given point $p_j$, it iterates through each point $p_k \in Dom_j^k$ (line 1) according to the dominance order and gets the nearest neighbours (line 2) to check whether the distance between the points in dominated set is less than the distance between $p_i$ and $p_j$ (line 3). 
Then, the points are added to the pruned point set (i.e., set of non-dominant points) according to Theorem~\ref{thm:theorem_3} (line 3-4) and Theorem~\ref{thm:theorem_4} (line 5-8). Finally, the set of non-dominant points is returned (line 9).

\begin{algorithm}[t]
	\begin{small}
	\KwData{An indoor partition $I$, A set of categories $\Psi$,}
	\KwResult{Update set of points in given partition $I$}
	
	$\overline{\mathcal{P}_i} \gets \emptyset$ , $\forall c_i \in \Psi$;\\
	\ForEach{$\{d_i,d_j\} \in   I(N)$ }{
		\ForEach{$\{c_a,c_b\} \in \Psi$}{
			$S_a,S_b \gets selectPoints(d_i,d_j,c_a,c_b)$
			$\overline{\mathcal{P}_a} \gets \overline{\mathcal{P}_a} \cup S_a$;\\
			$\overline{\mathcal{P}_b} \gets \overline{\mathcal{P}_a} \cup S_b$;\\
		}		
	}			
	\tcp{Update set of points in indoor partition $I$}					
	$\mathcal{P}_i = \overline{\mathcal{P}_i}$ , $\forall c_i \in \Psi$;\\
	\end{small}
	\caption{Dominant Based Pruning Algorithm}
\label{algo:dom_filter}
\end{algorithm}

Next, we present the dominance-based pruning algorithm that reduces the number of points in an indoor partition by eliminating non-dominant points belong to a given set of categories. 
As Algorithm ~\ref{algo:dom_filter} illustrates, for a given set of categories $\Psi$, the dominance-based pruning algorithm determines the set of dominant points for each given category using Algorithm~\ref{algo:select_points} (line 4) and updates the set of points of indoor partition $I$ accordingly (line 7). Since our pruning techniques are based on pairs of doors and categories,  we consider all possible combinations of doors and given categories to preserve the correctness of the algorithm. 
For example, let $I$ be an indoor partition consists of two doors $d_{10},d_{20}$ where sets of indoor points of $I$ for given categories $c_1, c_2$ are as follows : $c_1=\{p_1, p_2, p_3, p_4, p_5\}$ and $c_2=\{p_6, p_7, p_8, p_9\}$.
The outer for loop of the algorithm runs four times corresponding to the number of door pairs, i.e., $\{d_{10},d_{10}\}, \{d_{10},d_{20}\}, \{d_{20},d_{10}\}, \{d_{20},d_{20}\}$. And the inner for loop runs only once since there is only two given categories. Therefore, Algorithm~\ref{algo:select_points} is executed four times and the selected points are returned in each iteration. Assume that we obtain the results as follows: $\{\{p_1,p_2\}, \{p_8\}\}$, $\{\{p_2,p_3\}, \{p_7,p_8\}\}$, $\{\{p_1\}, \{p_8\}\}$ and $\{\{p_2\}, \{p_7\}\}$. Then, the set of points of indoor partition $I$ is updated as follows : $c_1=\{p_1,p_2,p_3\}$ and $c_2=\{p_7,p_8\}$. The indoor points $p_4, p_5$ and $p_6, p_9$ are eliminated.
\subsection{\algoAbb-dom Algorithm}\label{sub_sec:gcnn_dom}
As we mentioned earlier, we can accelerate the performance of \algoName~(\algoAbb) algorithm by reducing the number of indoor points accessed by the algorithm in query processing. Hence, we introduce an improved version of \algoAbb~algortihm denoted by \algoAbb-dom that utilize aforementioned dominance-based pruning technique to reduce the number of points in the indoor space. \algoAbb-dom algorithm is supported by a pre-processing approach in which we iterate the collection of indoor partitions covering a selected set of categories and eliminate all the non-dominant points by utilizing Algorithm~\ref{algo:dom_filter}.
Then we update the inverted VIP-tree by removing all non-dominant points of each selected categories while retaining the dominant indoor points. Also, we maintain
the minimum static score summary of each tree node updated accordingly.

Note that, the effectiveness of the pre-processing approach depends on the selected set of categories. Because, more incompetent points can be pruned only if the categories that we select, have a large number of points in the indoor space.
Therefore, we identify a set of categories that are more frequent in real-world queries.  
Intuitively, the most frequent query categories are the ones that have a higher demand.
Therefore, the indoor points (i.e., objects) belongs to these categories are highly available in the indoor venue. 
We show the effectiveness and the accuracy of the pre-processing approach in our empirical study. 
\section{Experiments} \label{sec:experiments}

\subsection{Experimental Settings}
\textbf{Indoor Venue and Category Datasets.} We use Chadstone Shopping Centre~\footnote{https://www.chadstone.com.au/} as our indoor venue.  
The Chadstone Shopping Centre which is the largest shopping centre in Australia currently features more than 300 retail outlets across 4 levels, with a total retail floor area over 200,000 $m^2$.
The floor plans of the Chadstone Shopping Centre is manually converted into machine-readable indoor venues. 
Coordinates of the buildings are obtained using OpenStreetMap~\footnote{https://www.openstreetmap.org/} and the sizes of indoor partitions (e.g. rooms, hallways) are determined.
Moreover, a three-dimensional coordinate system is used to represent an indoor entity in the dataset, in which the first two dimensions represent x and y coordinates of the indoor entry while the third represents the floor number. 
The relevant D2D graph consists of 339 vertices and 3867 edges.

We crawled data from the websites of major supermarkets (e.g., Coles, Woolworths, \textit{etc.}) as well as major retail stores (e.g., JB Hi-fi, Big W, \textit{etc.}) and obtained 140,000 objects along with their categories such as dairy, pantry, \textit{etc}. Then, each object was mapped into the particular indoor partition (e.g., a retail store) by randomly determining the location of the object inside the partition. 
Moreover, we obtained a larger dataset by replicating the real-world dataset four times (denoted by $REP$). Each object is replicated and randomly relocated with the same the category.\\
\hfill
\linebreak
\textbf{Query Generation.} 
We generated 5 query sets per dataset to study the performance of the algorithms. In order to generate query sets,
we took into account a property called objects per category (denoted by $\Omega$) which is 
the number of objects in the indoor space that belongs to a category. 
First, we identified five category sets w.r.t the aforementioned property, namely XS, S, M, L and XL. The category set XS was obtained by selecting the categories that have 80 - 120 objects in the indoor space. Similarly, the other category sets were obtained from  450 - 550, 950-1050, 1450-1550 and 1950 - 2050 objects respectively.
A query is generated by randomly selecting categories from the corresponding category sets and randomly determining the source and target points in the indoor space. Accordingly, 
50 queries were generated for each category set.
Moreover, we followed the same procedure to obtain different query sets for $REP$ dataset.  
\\
\hfill
\linebreak
\textbf{Competitors.} 
We compare our proposed algorithms, i.e., \algoAbb-dom and \algoAbb~, 
with an extension of an algorithm called \textit{global minimum path} (GMP)~\cite{yao2011multi}
which is the state of the art algorithm for outdoor space that solves a similar problem.
We extended their algorithm (denoted by iGMP) to support the problem of \queryAbb~by including additional attributes in the route search. Moreover, 
we utilized the state-of-the-art indoor indexing called VIP-tree~\cite{shao2016vip} with an inverted file in implementing the iGMP algorithm. Thus, it supports both efficient indoor distance computation and category-based filtering. 

\hfill
\linebreak
\textbf{Setup.}
In the real-world dataset, we observed that the categories which have a large number of objects in the indoor space, e.g., the categories in category set XL, are more clustered while the categories in category set XS, S, M are well distributed in the indoor space. Hence, as we explained earlier, we replicated the real-world dataset and obtained the $REP$ dataset. And, we take into account the same object ranges, i.e., 80- 120, 450 - 550, 950-1050, 1450-1550 and 1950 - 2050, when we were selecting categories for the category sets. This allowed us to select categories per category set that are well distributed in indoor space. Thus, 
the query categories of $REP$ dataset are well distributed and have a higher object density compared to the query categories of real-world dataset. 

Table~\ref{tab:default_settings} shows  the default settings we used in our experiments. The percentage ($\Delta$) denotes the percentage of query categories pre-processed.  
Suppose an approximate method $X$ returns a route $R$ for \queryAbb~query where the optimal route is $R^{opt}$ for the same instance. Then, $X$'s approximation ratio $r = cost(R) / cost(R^{opt})$. Moreover, 
All algorithms were implemented in C++ and our experiments were conducted on a Linux platform running on an Intel Core i5 $@$ 3.30GHz and 4GB RAM.
\begin{table}[t]
\centering
\caption{The parameters used for experiments}
\label{tab:default_settings}       
\begin{tabular}{|l|l|l|}
\hline
Parameter & Default & Range  \\
\hline
		Objects per category ($\Omega$) & M & XS, S, M, L, XL\\
		\hline
		Query categories ($q.\psi$) & 6 & 2, 4, 6, 8, 10 \\
		\hline
		Preference parameter ($\alpha$) & 0.5 & 0.1, 0.3, 0.5, 0.7, 0.9 \\
		\hline
		Percentage ($\Delta$) & 50 & 100, 80, 50, 20, 10, 0 \\
\hline
\end{tabular}
\end{table}
\subsection{Experimental Results}
In all experiments, we use the default settings while varying a single parameter at a time. Moreover, we report the average runtime in milliseconds and the approximation ratio for each experiment.
\subsubsection{Query Performance} 
First, we investigate the performance of algorithms on the real-world dataset. 
Under the default settings, each query consists of 6 categories in which each category has around  1000 related objects. 
Even though there are only around 6000 related objects in the indoor space, each algorithm deals with 140,000 objects in query processing. However, all the algorithms (including iGMP) perform efficiently since they support efficient category-based filtering and indoor distance computation.
\begin{figure}[t]
	\subfloat[]{\includegraphics[width=0.23\textwidth]{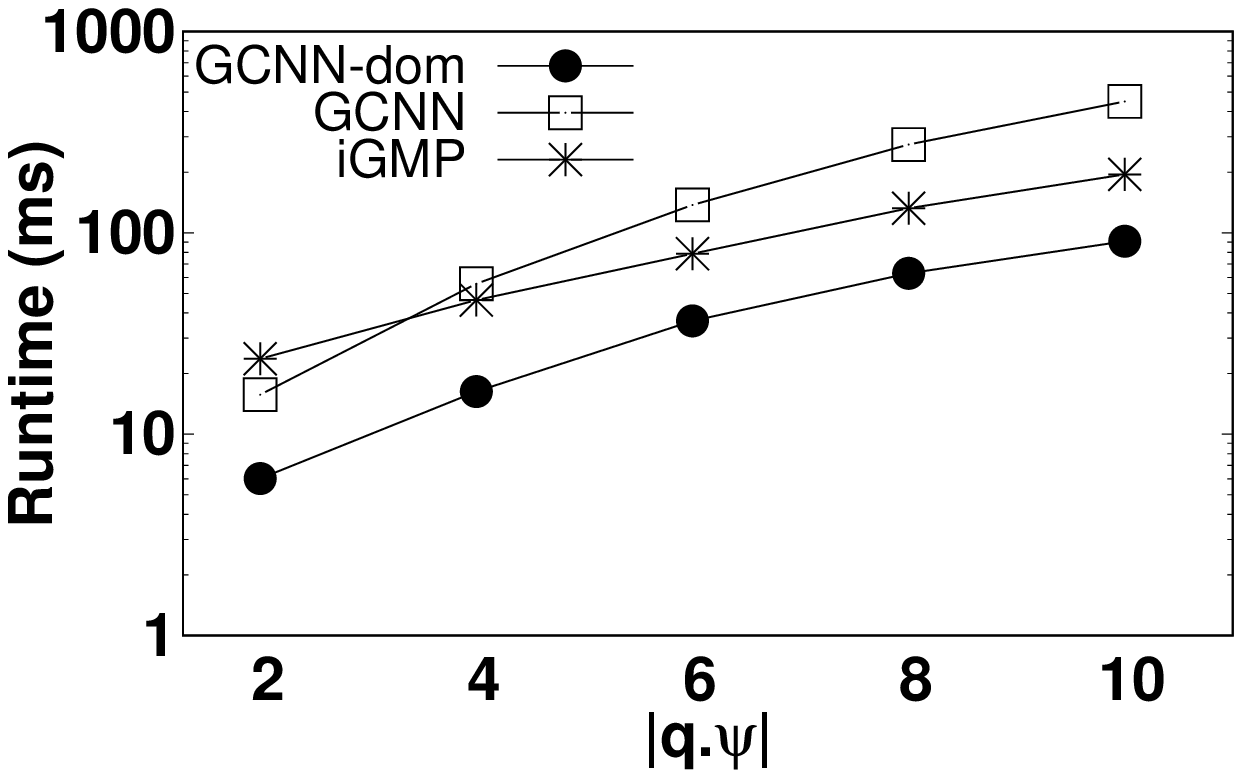}}
	\hspace{0.5mm}  
	\subfloat[]{\includegraphics[width=0.23\textwidth]{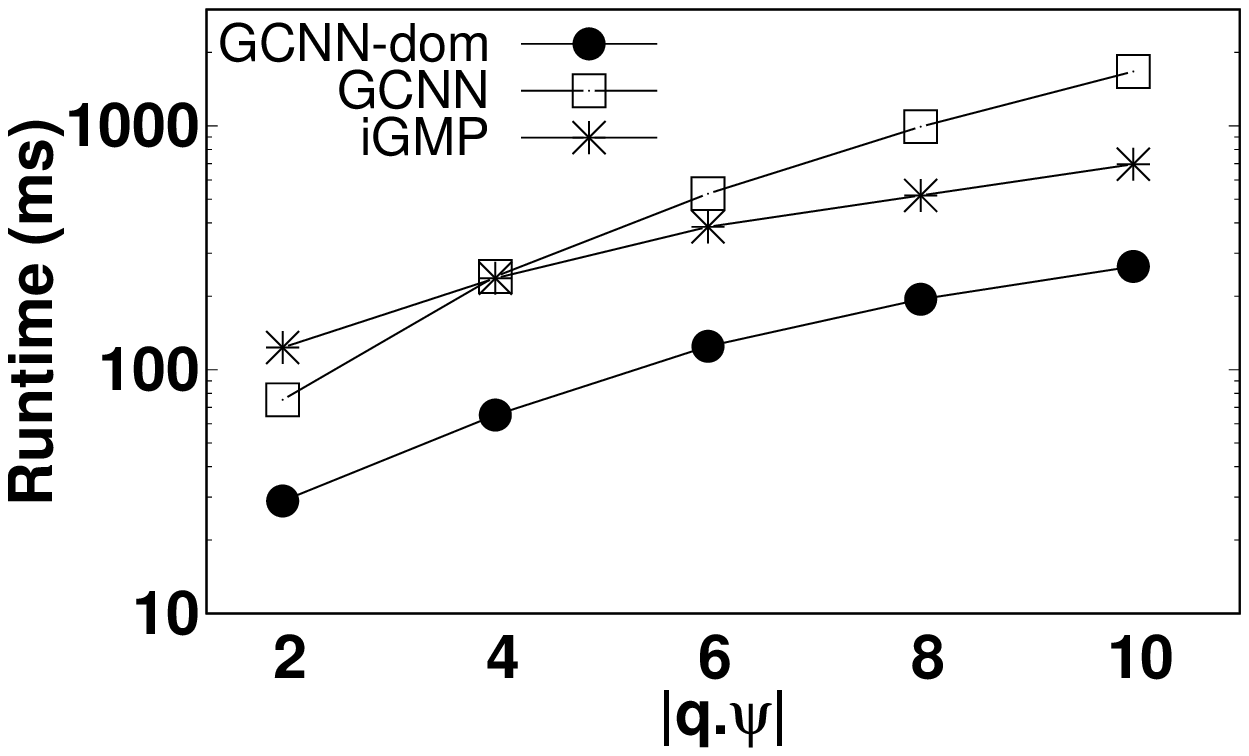}}
	\\
	\subfloat[]{\includegraphics[width=0.23\textwidth]{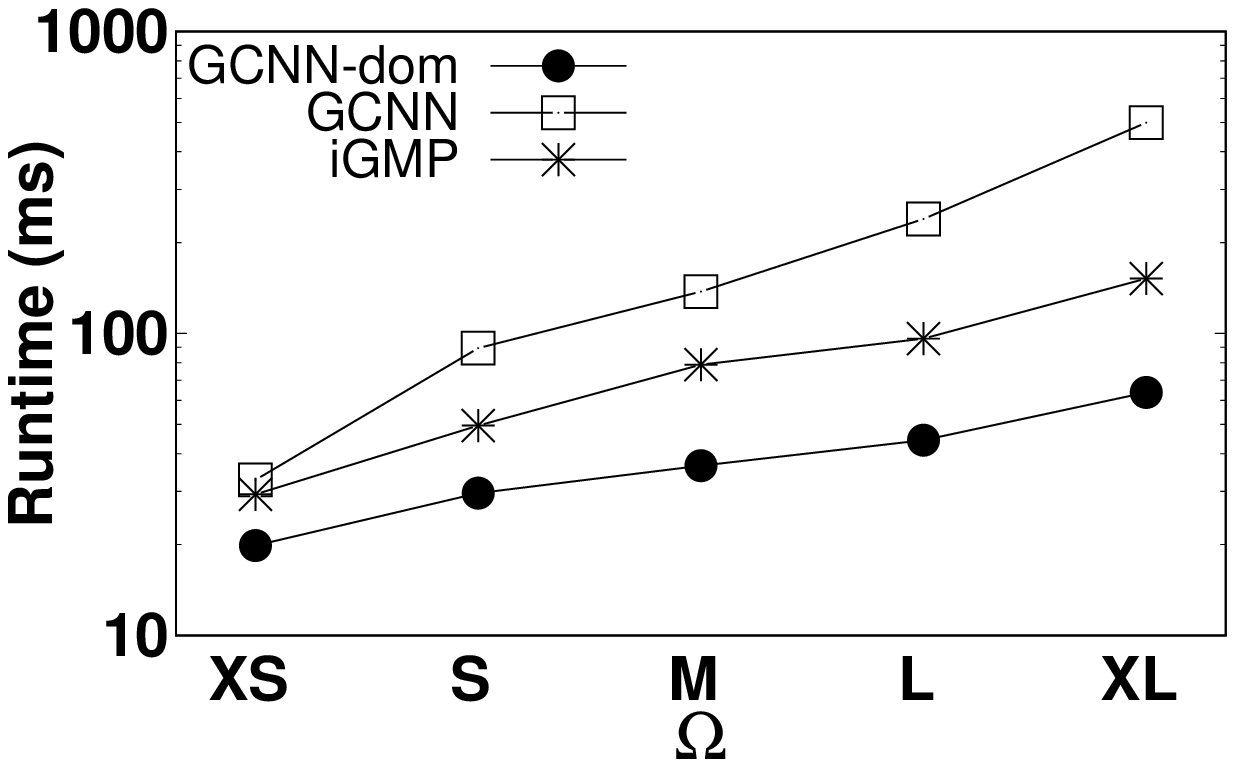}}	
	\hspace{0.5mm} 
	\subfloat[]{\includegraphics[width=0.23\textwidth]{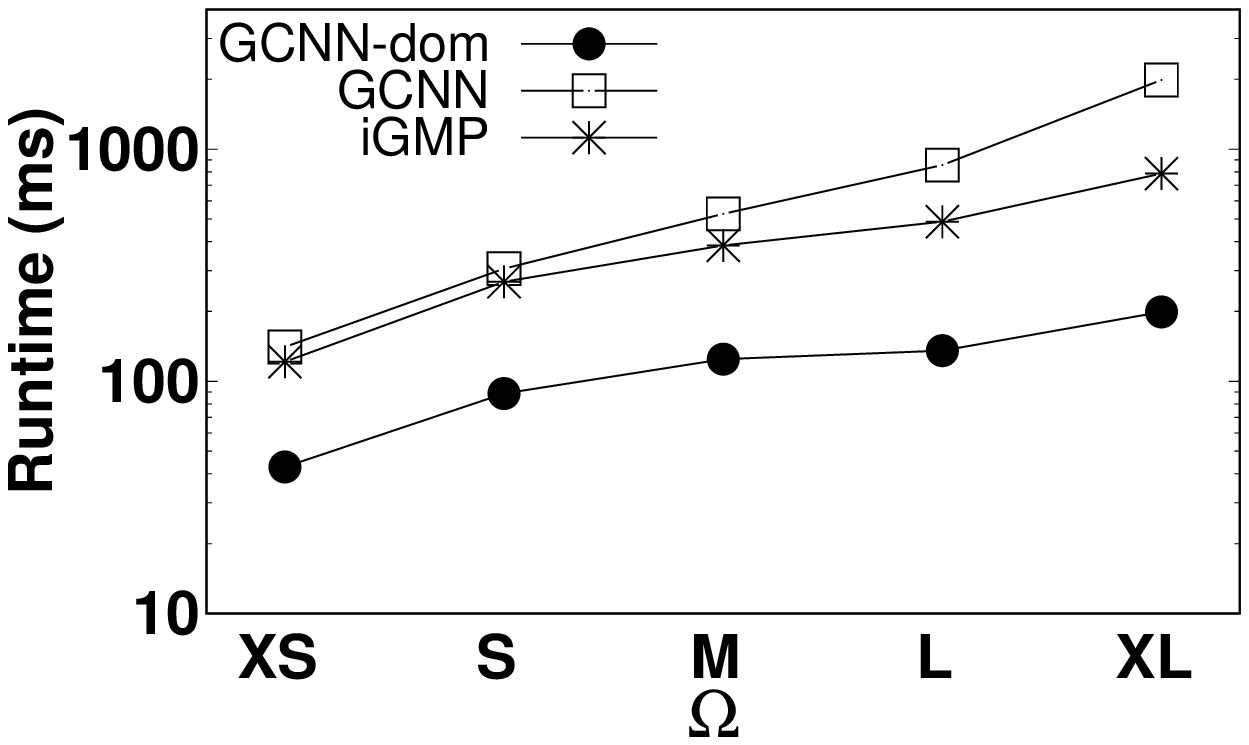}}
	
	\caption{Query response time on real-world and $REP$ datasets}
	\label{fig:exp_results_1}       
	\vspace{2mm}
\end{figure}
Figure~\ref{fig:exp_results_1}(a) reports the run time of the algorithms when we vary the number of query categories. 
The runtime of all algorithms increases when $|q.\psi|$ is increased as many ranking operations are carried out in query processing. As we explained in Section~\ref{sub_sec:cnn}, the number of query categories has a significant impact on the runtime of the algorithms.
The runtime of iGMP rises steadily as opposed to our algorithms since the related objects are ranked only once before it start constructing the route for a given query. 
Although \algoAbb~ranks indoor objects multiple times, it is reasonably efficient compared to iGMP since it utilizes the inverted VIP-tree which supports simultaneous travel and static costs based filtering. 
The runtime of \algoAbb-dom is generally 5-6 times better than \algoAbb. This is because \algoAbb-dom ranks a less number of objects as it pre-processes the dataset and eliminates non-dominant objects. 
Moreover, \algoAbb-dom answers a \queryAbb~query less than 0.1 seconds while iGMP is 2 times worse when $q.\psi = 10$.

Figure~\ref{fig:exp_results_1}(c) shows the runtime of all algorithms when we vary the objects per category, i.e., number of indoor objects per query category. Since each query consists of 6 categories under the default settings, the average number of related objects in the indoor space is 0.6K, 3K, 6K, 9K and 12K respectively. Clearly, the runtime of all algorithms increase as expected. We can see that \algoAbb~becomes much worse as $\Omega$ is increased. The reason is, \algoAbb~has to carry out more ranking operations when the related objects in the indoor space increase. \algoAbb-dom outperforms \algoAbb~by an order of magnitude while iGMP is 2.5 times slower than \algoAbb-dom when $\Omega = 2000$. 

Figure~\ref{fig:exp_results_1}(b) and Figure~\ref{fig:exp_results_1}(d) investigate the performance of all algorithms on $REP$ dataset.
As we mentioned earlier, the query categories are well distributed on $REP$ dataset. Hence, the number of indoor partitions that cover the query categories are higher compared to the real-world dataset. 
Thus, the runtime of algorithms is increased since the ranking operations become expensive.
As Figure~\ref{fig:exp_results_1}(b) shows the runtime of \algoAbb~is almost 2 seconds when $|q.\psi| = 10$. But, \algoAbb-dom takes only 0.2 seconds while iGMP is 4 times slower. According to Figure~\ref{fig:exp_results_1}(d), clearly, \algoAbb-dom outperforms all other algorithms as it takes only 0.2 seconds to answer a \queryAbb~query when $\Omega = 2000$.  
Distinctly, \algoAbb-dom is superior to all other algorithms as it uses the dominance-based pruning technique in pre-processing. The results conclude that the dominance-based pruning technique is much effective in accelerating the performance of the proposed algorithm.  
\begin{figure}[t]
	\subfloat[]{\includegraphics[width=0.23\textwidth]{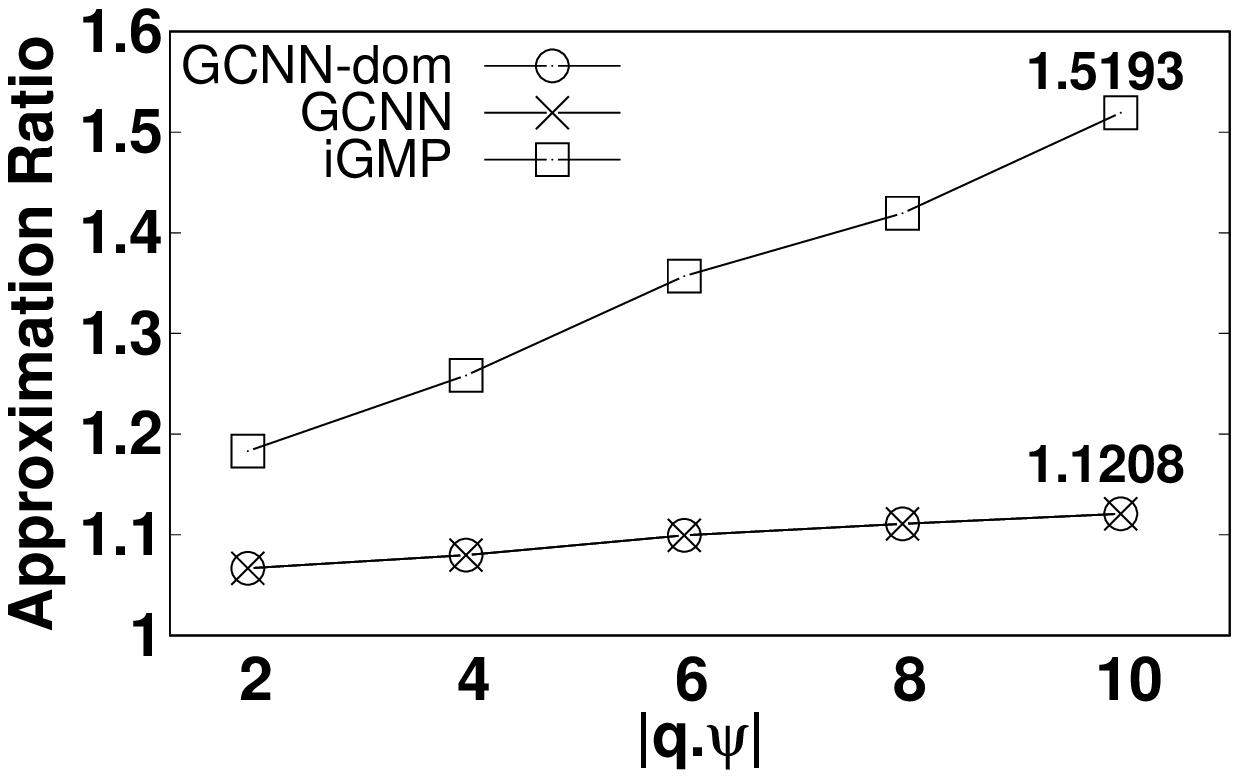}}
	\hspace{0.5mm}  
	\subfloat[]{\includegraphics[width=0.23\textwidth]{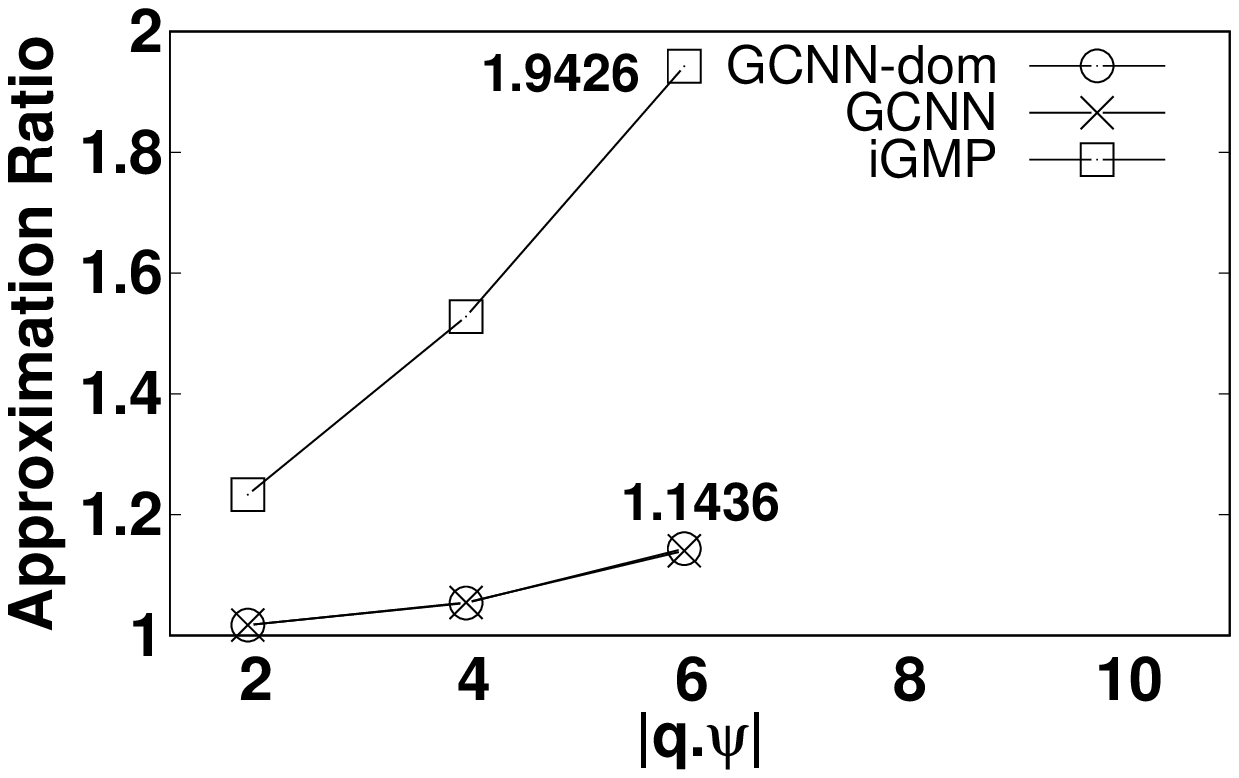}}
	\\
	\subfloat[]{\includegraphics[width=0.23\textwidth]{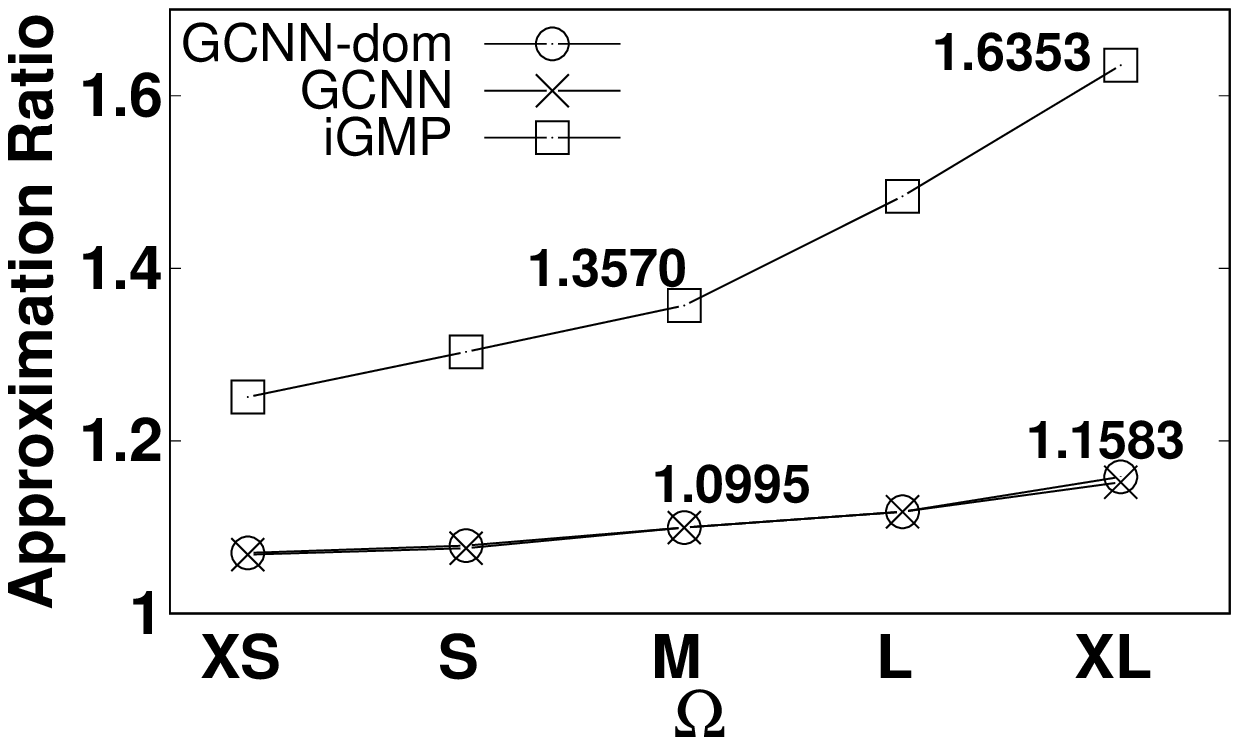}}
	\hspace{0.5mm} 
	\subfloat[]{\includegraphics[width=0.23\textwidth]{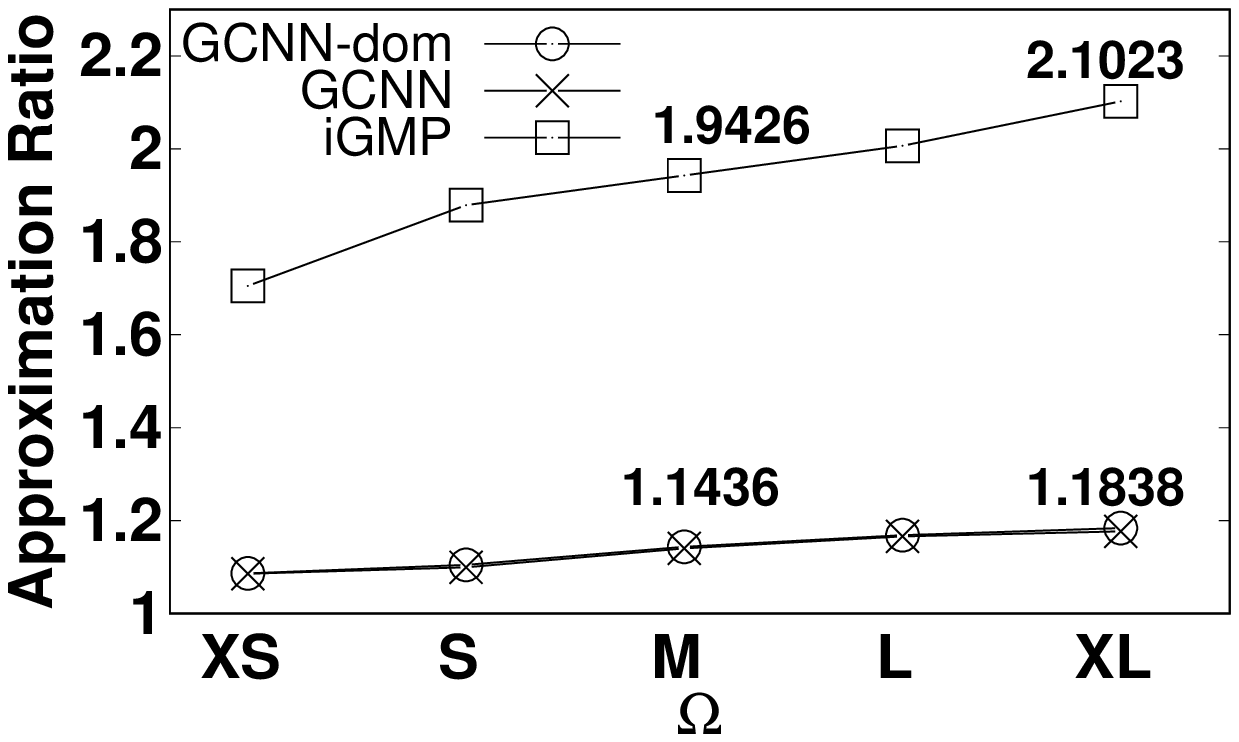}}
	\\
	\subfloat[]{\includegraphics[width=0.23\textwidth]{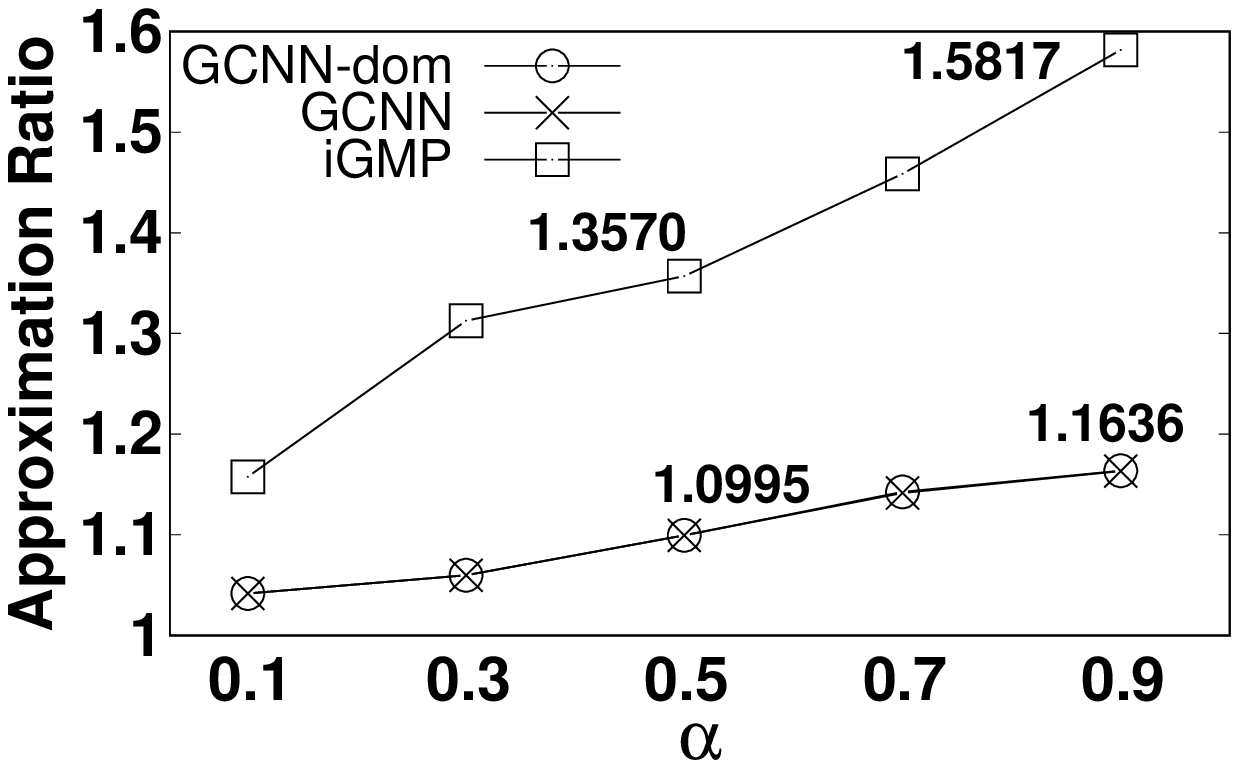}}
	\hspace{0.5mm}  
	\subfloat[]{\includegraphics[width=0.23\textwidth]{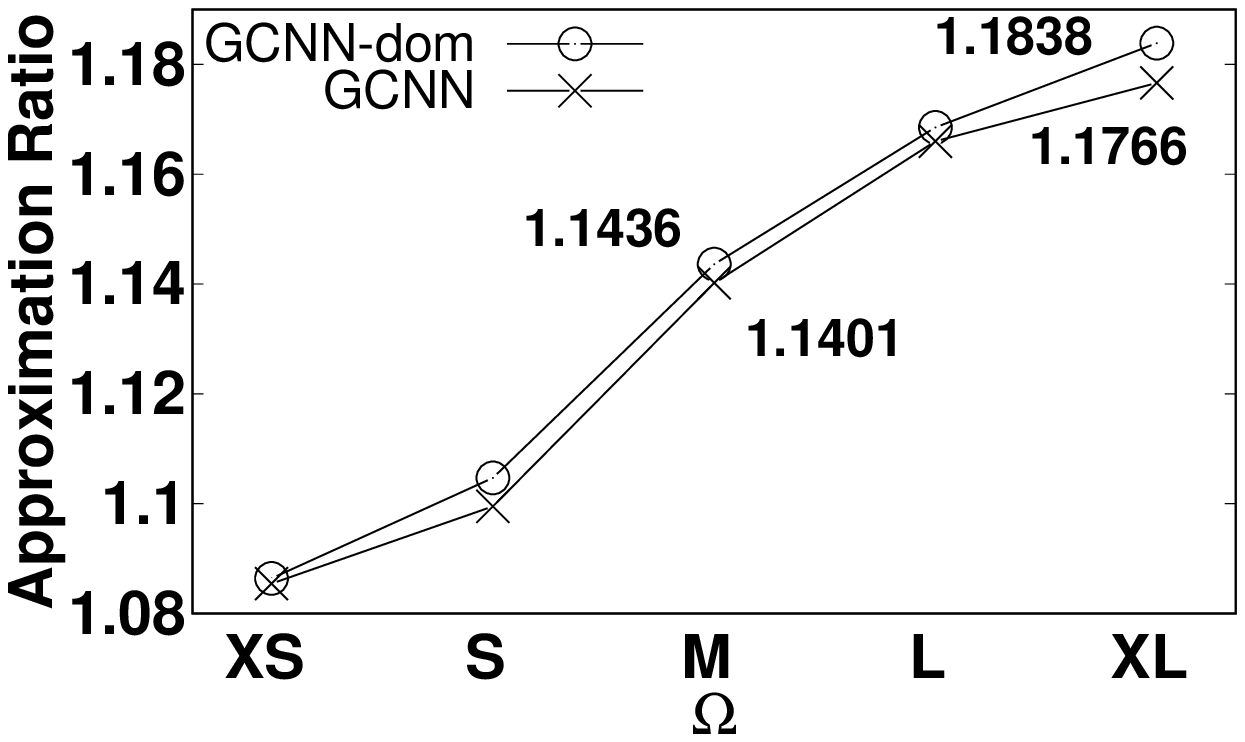}}
	\caption{Approximation quality on real-world and $REP$ }
	\label{fig:exp_results_2}       
	\vspace{2mm}
\end{figure}
\subsubsection{Accuracy of Approximations}
The objective of this set of experiments is to study the accuracy of the approximation algorithms. Note that, the results for some of the settings are unavailable since the brute force algorithm failed to finish after a reasonable time. 
Figure~\ref{fig:exp_results_2}(a) to Figure~\ref{fig:exp_results_2}(d) report the approximation ratio of algorithms for the corresponding experiments in Figure~\ref{fig:exp_results_1}. As Figure~\ref{fig:exp_results_2}(a) depicts,
iGMP has the worse approximation quality for all the different settings of $|q.\psi|$. When $|q.\psi| = 10$, the approximation ratio of iGMP is slightly higher than $1.5$ while both \algoAbb-dom and \algoAbb~are around $1.1$.  
According to the Figure~\ref{fig:exp_results_2}(c) approximation ratio of all algorithms increases when we increase the $\Omega$. We can see that the approximation ratio of iGMP increases drastically after $\omega = 1000$. This is because of the category distribution. The approximation ratio of iGMP is 1.6 when $\Omega = 2000$, while our algorithms stay close 1.1 in all cases. Clearly, the approximation ratio of our algorithms are almost similar in all cases. 
Figure~\ref{fig:exp_results_2}(e) shows the approximation ratio of algorithms when we vary the query preference parameter. The approximation ratio of all algorithms affected by the alpha value. All algorithms have the worse approximation ratio when $\alpha = 0.9$. The approximation ratio of iGMP is close 1.6 while our algorithms are under 1.2. Note that, the approximation ratio of \algoAbb-dom and \algoAbb~are almost same. 
Next, we examine the accuracy of approximations on $REP$ dataset. Figure~\ref{fig:exp_results_2}(b) shows that the ratio of iGMP is worse in all cases where it is closer to 2 even when $|q.\psi| = 6$. But, \algoAbb-dom and \algoAbb~stays under 1.2. We can see that when the distribution of the categories change, the approximation ratio of iGMP drastically increases. As Figure~\ref{fig:exp_results_2}(d) depicts the ratio of iGMP exceeds 2 for the large $\Omega$ values. The approximation ratio of our algorithms show much better approximation quality by being consistent around 1.1 for the large dataset.

Finally, we compare the approximation quality of our algorithms on $REP$ dataset. Here, we have excluded the results for real-world dataset since the approximation ratios of both algorithms almost similar. 
Figure~\ref{fig:exp_results_2}(f) shows the difference between the approximation ratio of our algorithms while varying the objects per category, i.e., $\Omega$. 
The approximation ratio of \algoAbb-dom has deviated from \algoAbb~by 0.01 when $\Omega = 2000$. For all other cases, clearly, the difference is negligible. This insignificant difference in ratios indicates the accuracy of dominance-based pruning technique in identifying the incompetent indoor points in the indoor space.

\subsubsection{Effect of pre-processing}
\begin{figure}[t]
	\centering
	\includegraphics[width=0.25\textwidth]{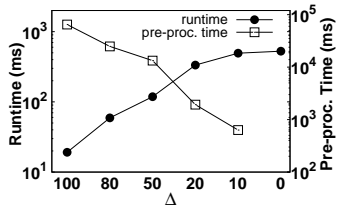}
	\caption{Effect of Pre-processing}
	\label{fig:exp_results_3}
	\vspace{2mm}
\end{figure}

Figure~\ref{fig:exp_results_3} reports the runtime of \algoAbb-dom and the corresponding pre-processing time while varying the percentage of query categories that has been pre-processed (i.e., $\Delta$). Note that, we denote the pre-processing time by \textit{pre-proc. time} in Figure~\ref{fig:exp_results_3}. 
The percentage of query categories $\Delta = 0$ indicates that no pre-processing is done and $\Delta = 100$ indicates that all the query categories are pre-processed. Thus, $\Delta = 50$ denotes that half of the query categories are identified as highly frequent categories for pre-processing.
As Figure~\ref{fig:exp_results_3} shows the runtime of \algoAbb-dom decreases as we pre-process more query categories. The reason is, for large $\Delta$ values, \algoAbb-dom ranks only a small set of objects as most of the incompetent objects were eliminated in pre-processing.
When $\Delta = 100$, \algoAbb-dom can answer a \queryAbb~query in 0.04 milliseconds 
in which it outperforms iGMP in an order of magnitude.  
The approximation quality increases as we decrease $\Delta$. But the variation is insignificant. For example, the difference between  approximation ratio of $\Delta =0$  and  $\Delta = 100$ is 0.01. Hence, we do not report the approximation ratios corresponding to the experiment in Figure~\ref{fig:exp_results_3}. The results conclude that our pre-processing approach is highly accurate and effective in accelerating the performance of the proposed algorithm.

\section{Conclusion} \label{sec:conclusion}
In this paper, we define the problem of \queryName~query, denoted by \queryAbb, which returns a route from a given source point to a target point that passes through at least one indoor point from each given category while minimizing the cost of the route in terms of travel and static costs. The problem of answering \queryAbb~query is NP-hard. 
Based on a novel dominance-based pruning, we devise an efficient approximation algorithm called \algoName~(\algoAbb) algorithm to answer \queryAbb~queries. The experimental results demonstrate that the proposed algorithm is highly efficient and offer high-quality results.

For future work, we plan to extend our algorithm to support indoor points with multiple keywords. Thus, the textual similarity will also be considered in route cost computation. 

\bibliographystyle{ACM-Reference-Format}
\bibliography{ms}

\end{document}